\documentclass[10pt,twocolumn,letterpaper]{article}

\usepackage{iccv}              % To produce the CAMERA-READY version
\definecolor{iccvblue}{rgb}{0.21,0.49,0.74}
\usepackage[pagebackref,breaklinks,colorlinks,citecolor=iccvblue]{hyperref}

\usepackage{url}            % simple URL typesetting
\usepackage{booktabs}       % professional-quality tables
\usepackage{nicefrac}       % compact symbols for 1/2, etc.
\usepackage{xcolor}         % colors
\usepackage{multirow}
\usepackage{bm}

\usepackage{amsmath}

\usepackage[ruled]{algorithm2e}
\usepackage{algpseudocode}
\usepackage{graphicx}  %Required
\usepackage{CJK}
\usepackage{bbm}
\usepackage{float}
\usepackage{xcolor,colortbl}

\usepackage{bigstrut,bigdelim}
\usepackage{diagbox}
\usepackage{wrapfig}
\usepackage{verbatim}
\usepackage{afterpage}
\usepackage{enumitem}

\usepackage{subcaption}
\usepackage{multicol}
\usepackage{multirow}
\usepackage{amssymb}
\usepackage{xcolor}
\usepackage{caption}
\usepackage[noabbrev,capitalize]{cleveref}
\usepackage{acronym}
\usepackage{xspace}
\usepackage{pifont}
\usepackage{lipsum}
\usepackage{makecell}
\usepackage{tablefootnote}
\usepackage{footnote}
\usepackage[normalem]{ulem}
\usepackage[misc]{ifsym}

\usepackage{listings}
\usepackage{titletoc}

\makeatletter
\DeclareRobustCommand\onedot{\futurelet\@let@token\@onedot}
\def\@onedot{\ifx\@let@token.\else.\null\fi\xspace}

\def\ie{\emph{i.e}\onedot}

\def\wrt{w.r.t\onedot}

\makeatother

\makeatletter
\renewcommand{\paragraph}{%
  \@startsection{paragraph}{4}%
  {\z@}{0ex \@plus 0ex \@minus 0ex}{-1em}%
  {\normalfont\normalsize\bfseries}%
}
\makeatother

\usepackage{scalerel}

\newcommand{\model}{VideoLLaMB\xspace}
\newcommand{\bench}{Needle In A Video Haystack\xspace}
\newcommand{\benchshort}{NIAVH\xspace}
\newcommand{\hlt}[1]{\cellcolor{blue!25}{#1}}
\newcommand{\hltse}[1]{\cellcolor{blue!15}{#1}}
\newcommand{\hlttrd}[1]{\cellcolor{blue!7}{#1}}

\definecolor{LightCyan}{rgb}{0.88,1,1}

\def\paperID{7824} % *** Enter the Paper ID here
\def\confName{ICCV}
\def\confYear{2025}

\title{VideoLLaMB: Long Streaming Video Understanding \\ with Recurrent Memory Bridges}

\author{\bf Yuxuan Wang$^{\,1}$\thanks{Equal contributions} \qquad Yiqi Song$^{\,1,2*}$ \qquad Cihang Xie$^{\,3}$ \qquad Yang Liu$^{\,4}$ \qquad Zilong Zheng$^{\,1}$\thanks{Correspondence to Zilong Zheng $\langle$zlzheng@bigai.ai$\rangle$.} \vspace{.05in}\\
$^{1}$ NLCo Lab, State Key Laboratory of General Artificial Intelligence, BIGAI\\
$^{2}$ School of Computer Science \& Technology, Beijing Institute of Technology\\
$^{3}$ Computer Science and Engineering, University of California\\
$^{4}$ Wangxuan Institute of Computer Technology, Peking University \\
{\tt\small flagwyx@gmail.com, yiqis@bit.edu.cn, zlzheng@bigai.ai} \vspace{.05in} \\
\url{https://github.com/bigai-nlco/VideoLLaMB}
}

\begin{document}
\maketitle

\begin{abstract}
Recent advancements in large-scale video-language models have shown significant potential for real-time planning and detailed interactions. However, their high computational demands and the scarcity of annotated datasets limit their practicality for academic researchers.
In this work, we introduce \textbf{VideoLLaMB}, a novel and efficient framework for long video understanding that leverages recurrent memory bridges and temporal memory tokens to enable seamless encoding of entire video sequences with preserved semantic continuity. Central to our approach is a \textit{SceneTiling} algorithm that segments videos into coherent semantic units, facilitating robust understanding across tasks without requiring additional training. VideoLLaMB achieves state-of-the-art performance, surpassing existing models by $4.2$ points on four VideoQA benchmarks and by $2.06$ points on egocentric planning tasks. Notably, it maintains strong performance under extreme video length scaling (up to $8\times$) and excels at fine-grained frame retrieval on our proposed \textbf{Needle in a Video Haystack (NIAVH)} benchmark. With linear GPU memory scaling, VideoLLaMB processes up to 320 frames using a single Nvidia A100 GPU, despite being trained on only 16 frames—offering an unprecedented balance of accuracy, scalability, and cost-effectiveness. This makes it highly accessible and practical for the academic community. \looseness=-1

\end{abstract}

% \vspace{-5mm}

\section{Introduction}\label{sec:intro}

Recent advancements in large-scale video language models, exemplified by GPT-4o
% \footnote{\url{https://openai.com/index/hello-gpt-4o/}} 
and Project Astra
% \footnote{\url{https://deepmind.google/technologies/gemini/project-astra/}},
have captivated global attention due to their potential for sophisticated interaction with real-world environments~\cite{jia2024langsuite,zheng2025mcu}. These models are particularly noteworthy for their capacity to comprehend streaming video~\cite{cvpr25omnimmi}, which can be conceptualized as video with an unlimited context length. This capability necessitates both the observation of the current state and the ability to leverage long-term memory. Despite their promise, the training of such large-scale video-language (VidL) foundational models remains impractical for academic researchers. This impracticality arises from the substantial computational resources required by the complex, high-dimensional nature of long streaming video data, in addition to the scarcity of well-annotated, publicly available video-language datasets. These factors present significant challenges to scaling video-language models to the extent observed with large language models (LLMs).

To circumvent these challenges, the community has witnessed a growing interest in developing computationally efficient multimodal large language models~(MLLMs).
Traditional methods resort to \textit{video compression} strategies, such as sampling~\citep{videollama, videollava}, aggregation~\citep{pllava}, semantic consolidation~\citep{moviechat}, and resampling~\citep{vista-llama, malmm}, in order to temporally reduce the length of the video. Yet, these methods often lead to the \textbf{loss of critical visual cues}, undermining the model's ability to capture essential cues. Other approaches~\citep{streaminglm,wang2023vstar} segment videos into shorter clips to mitigate the computational load of processing long videos. However, segmentation can \textbf{disrupt the semantic flow of content}, complicating the encoding process and potentially impacting the general understanding of the video narrative. Lastly, 
prevalent video understanding benchmarks, primarily based on linguistic question-answering pairs, exhibit \textbf{static}~\citep{singularity} and/or \textbf{language biases}~\citep{languagebia-1,languagebias-2}. These biases favor models that rely more on static imagery or textual elements, respectively, and fail to provide a comprehensive assessment of a model's capability on extended video sequences. 
\looseness=-1

% Refer to $\S$\ref{sec:related} for detailed discussions.

% current video understanding benchmarks are primarily based on linguistic question-answering pairs, which, however, can introduce \textbf{static biases}~\citep{singularity} that favor models relying on static images over video sequences, and/or \textbf{language biases}~\citep{languagebia-1,languagebias-2} where the model's performance is unduly swayed by the textual elements, rendering them insufficient for effectively assessing a model's capacity for long video understanding; refer to $\S$\ref{sec:relate} for detailed discussion.

\begin{figure*}[t]
    \centering
    % \vspace{-1.5em}
    \includegraphics[width=.96\textwidth]{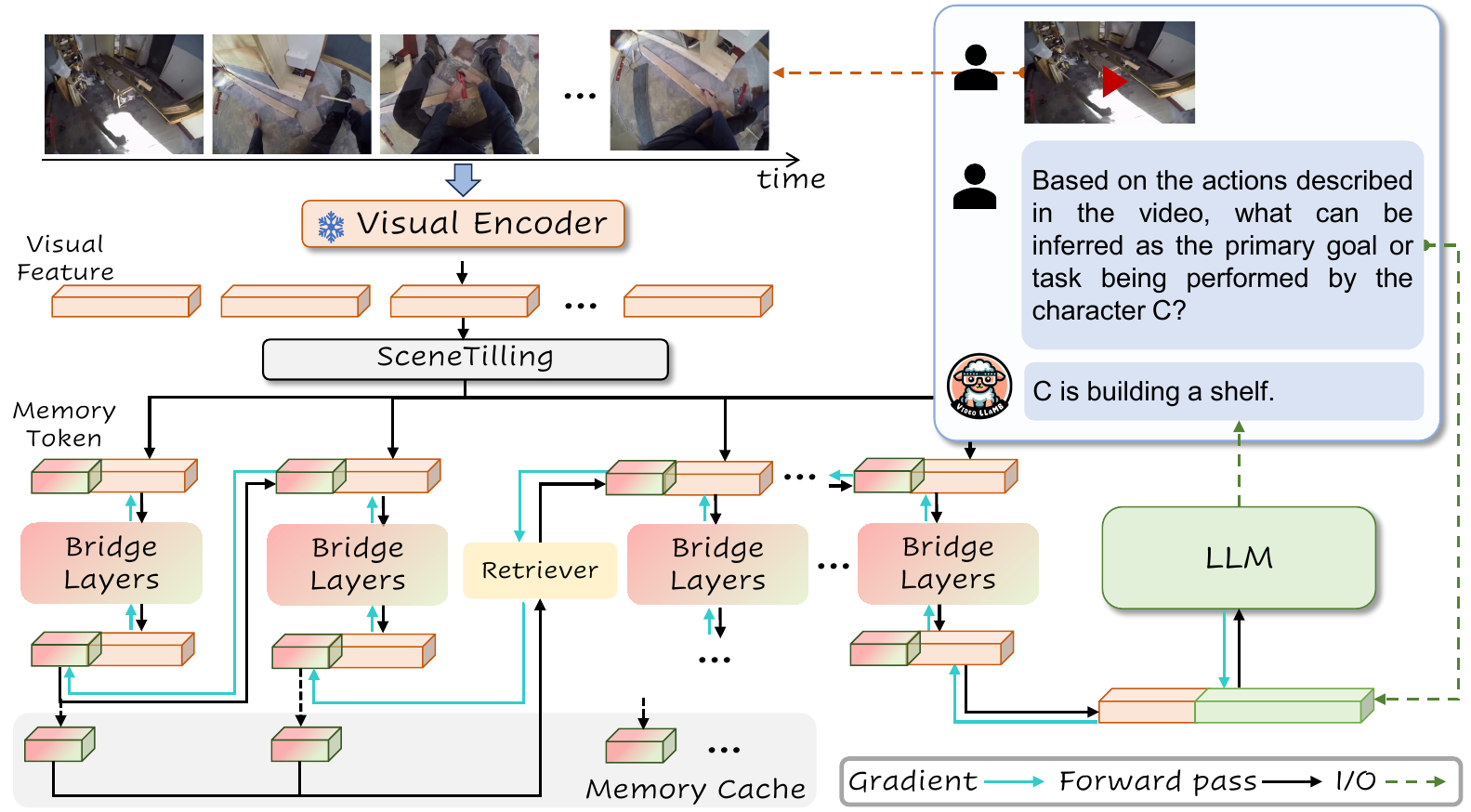}
    \caption{\textbf{Overview of \model.} We first extract the video features using an off-the-shelf vision encoder, then apply SceneTiling to segment the video into semantic segments ($\S$\ref{sec:segment}). Next, we use recurrent memory on these semantic segments to store video information within memory tokens ($\S$\ref{sec:recurrent}). We further employ a retrieval mechanism to update the memory tokens and address long-dependency issues($\S$\ref{sec:retrieval}). Finally, we project the memory-token-augmented features from the current video segment into the LLM.}
    \label{fig:overview}
    % \vspace{-.1in}
\end{figure*}

To address these multifaceted limitations, we introduce \textbf{\model}, an innovative framework that learns temporal \textbf{M}emory tokens within \textbf{B}ridge layers that recursively encode the entire video content, ensuring that no visual information is discarded deliberately (\Cref{fig:overview}; $\S$\ref{sec:method}). Specifically, we devise Memory Bridge Layers, equipped with recurrent memory tokens, that function without altering the architecture of the visual encoder and LLM. Furthermore, to mitigate the backpropagation through time (BPTT) issue, 
we maintain long-term dependencies by preserving recurrent memory tokens in a memory cache, which is periodically refreshed through a retrieval process. To compensate for the limitations of the sliding window technique, we propose the SceneTiling algorithm that divides the video into relatively independent sequences of semantic segments. This reduces the dimensions within each semantic unit without sacrificing semantic details. By constructing our recurrent memory with a retrieval mechanism based on these semantic segments, our method strikes a balance between effective and efficient comprehension of the current state and long-term memory retention.

% To benchmark the multimodal retrieval capabilities of long videos, in $\S$\ref{sec:bench}, we launch a pressure test called ``Needle in a Video Haystack''~(NIAVH), a synthetic benchmark that ensures an unbiased evaluation of long video understanding.
% NIAVH requires the model to answer questions related to the key information scattered throughout the video context. 
% NIAVH supports the identification of ``needles'' information in text, image, and video modality within extensive video content. This benchmark is designed to evaluate the ability of models to comprehend lengthy videos in a more direct, fair and comprehensive means to benchmark long video understanding across different dimensions. 

In $\S$\ref{sec:exp}, we highlight the empirical advantages of \model in comparison with prior arts as:
\begin{itemize}[leftmargin=*, noitemsep, topsep=0pt]
% \begin{itemize}
    % \item \textbf{Comprehensive long video understanding.} We demonstrate the effectiveness of \model using two long-form video QA benchmarks: EgoSchema~\citep{egoschema} and NexTQA~\citep{nextqa}. Our results show an average improvement of $5.5$ accuracy points over PLLaVA~\citep{pllava}, a model with the same initialization and training video dataset. Furthermore, \model maintains its performance even when the video length extends to $8$ times longer than the original. Additionally, performance on MVBench~\citep{mvbench} indicates that \model significantly outperforms prior models like PLLaVA using the same training data and LLM baseline.
    % \item \textbf{Enhanced frame retrieval in long videos.} To evaluate our model's ability in frame retrieval for long videos, we propose a multimodal Needle in a Video Haystack (NIAVH) test. This test requires the model to predict the true answer about an inserted image in a long video. In our NIAVH pressure test, which ranges from $1$ to $320$ seconds in length, \model consistently retrieves the correct image needles at various depths, outperforming other methods as video length increases.
    \item \textbf{Long-Term Memory Reservation.} We demonstrate the effectiveness of \model in comprehensive long video understanding and enhanced frame retrieval through rigorous testing on established benchmarks. Using VideoMME~\citep{videomme}, EgoSchema~\citep{egoschema} and NexTQA~\citep{nextqa}, \model shows an average improvement of $4.2$ accuracy points over PLLaVA~\citep{pllava}, despite utilizing the same initialization and training video dataset. Notably, \model maintains robust performance even when video lengths extend to eight times their original duration. To further evaluate frame retrieval capabilities, we introduce the multimodal Needle in a Video Haystack (NIAVH) test, where \model consistently identifies the correct image within videos ranging from 1 to 320 seconds, surpassing other methods as video length increases.
    \item \textbf{Real-time egocentric planning.} To evaluate our model's performance in video planning tasks, we used the dataset EgoPlan~\citep{egoplan}. Our method achieves the best performance among all 7B video-language models, showing an improvement of $2.06$ accuracy points over PLLaVA.
    \item \textbf{Training-free streaming captioning.} By employing the SceneTiling algorithm, our method can automatically predict the end of a caption in streaming video without relying on special tokens during the training phase.

\end{itemize}

\section{\model}\label{sec:method}

\model is an extensible framework designed to enhance long video understanding, composed of three key modules: semantic-based segmenter ($\S$\ref{sec:segment}), recurrent memory layer ($\S$\ref{sec:recurrent}), and memory retriever ($\S$\ref{sec:retrieval}). Each of these components will be detailed in the subsequent sections. \Cref{fig:overview} depicts the overall framework.

% \paragraph{Problem Formulation}
% Given a video $V$ consisting of a series of $n$ frames as $\{f_1, f_2, \ldots, f_n\}$ and a linguistic input referred to as $L$, the objective of video-language models is to produce a response that corresponds coherently with the context provided by both the video and the language input.

\subsection{SceneTiling: Segmentation with Semantics}
\label{sec:segment}
% , both in text~\citep{Song2016DialogueSS, Xu2021TopicAwareMD, Xing2021ImprovingUD} and video~\citep{Rao2020ALA,Chen2021ShotCS,Mun2022BoundaryawareSL,movienet,vstar},
Semantic segmentation along temporal sequence has long been recognized as an important task because it preserves the non-linear structure of context and greatly aids in compressing extensive context~\citep{Rao2020ALA,Chen2021ShotCS,Mun2022BoundaryawareSL,movienet,vstar}. 
To address the disruption of semantic flow (see $\S$\ref{sec:intro}), we introduce SceneTiling, a model-free scene segmentation algorithm inspired by TextTiling~\citep{DBLP:journals/coling/Hearst97}. SceneTiling divides the entire video sequence into segments that are semantically distinct, ensuring intra-segment coherence.
% To above disruption of semantic flow (refer to $\S$\ref{sec-intro}), in this work, inspired by TextTilling~\citep{DBLP:journals/coling/Hearst97}, 
% we apply video scene segmentation technology to divide our entire video into a series of semantically less relevant segments. Specifically, 
% we introduce SceneTiling, a model-free scene segmentation algorithm, to divide the entire video sequence into video segments such that each segment is semantically non-overlap with others, \ie, inter-segment coherence. 

Formally, given a sequence of $n$ frames $\{v_1, v_2, \ldots, v_n\}$,
the SceneTiling algorithm is as follows.
\begin{enumerate}[leftmargin=*, noitemsep, topsep=0pt]
    \item Compute the cosine similarity $S_C(\cdot, \cdot)$ between adjacent frame pairs using the \texttt{[CLS]} token from ViT, resulting in a sequence of similarity scores $\{c_1, c_2, \ldots, c_{n-1}\}$, where $c_i = S_C({\rm ViT}(v_i), {\rm ViT}(v_{i+1}))$. 
    \item Calculate the depth score for each point as $d_i = \left(cl_i+cr_i-2c_i\right)/{2}$, where $cl_i$ and $cr_i$ are the highest score to the left and right of $c_i$, respectively. A higher depth score indicates that the surrounding similarity is greater than at the point itself.
    \item Calculate the expectation $\mu$ and variance $\sigma$ of the depth scores $\{d_1, d_2, \ldots, d_{n-1}\}$. Set the segmentation threshold as $\mu + \alpha \cdot \sigma$, where $\alpha$ is a hyperparameter controlling the likelihood of segmenting the video. Select the $K-1$ depth scores that exceed the threshold to divide the video into $K$ semantic segments $\{s_1, s_2, \ldots, s_K\}$. Each segment represents a relatively independent semantic unit consisting of a sequence of frames.
\end{enumerate}

Aside from temporal semantic segmentation, SceneTiling enables streaming video captioning without requiring training with special tokens~\citep{video-online, streamingcap, vita}.
% as shown in \Cref{fig:streaming}).
% Additional details on streaming dense video captioning are provided in \textbf{\Cref{supp:streaming}}.

\subsection{Recurrent Memory Bridge Layers}
\label{sec:recurrent}
Traditional recurrent memory-based Transformers~\citep{rmt, rmt100m, rmt-r} incur significant computational costs when scaled up, \ie, $\mathcal{O}(L^2/K)$, where $L$ is the context length and $K$ is the number of segments, primarily due to its recurrent mechanism over the whole language model.
More recently, some works empirically identify that linear projection best withstands visual information within MLLMs~\citep{llava,llava15,llavanext}, albeit with high space complexity, whereas the resampler has strong compressing ability on semantic information~\citep{blip2}, though it tends to miss detailed information~\citep{slot}.

In this work, we devised a novel Recurrent Memory Bridge Layer, implemented as Transformer blocks, that integrates recurrent memory tokens within bridge layers to enhance the linear layer's memorization ability. 
Formally, for each video segment $s_i$, we prepend a fixed number of memory tokens, denoted as $[m_i; s_i]$, where $m_i$ represents the memory tokens. Subsequently, we apply standard self-attention to this sequence, yielding $[m_{i+1}; o_{i}] = {\rm BridgeLayer}([m_i; s_i])$. Here, $m_{i+1}$ is the updated memory token, and $o_{i}$ is the visual representation from the bridge layers. 
This process is carried out recursively, traversing the semantic video segments while updating the memory tokens. After a total of $k$ steps, this output represents the condensed visual summary of the video sequence and will serve as the input for the LLM.
As such, the Memory Bridge can compress past video into memory while preserving current video scenes through projection without losing detailed information. \looseness=-1

\subsection{Memory Cache with Retrieval}
\label{sec:retrieval}

One of the primary challenges associated with recurrent memory bridge layers is the potential for gradient vanishing, which can impede the model's ability to learn long-range dependencies~\cite{li2024loogle}. To mitigate this issue, we propose the incorporation of a memory cache with a retrieval strategy designed to preserve previous states of memory. 

\paragraph{Memory Attention} At each timestep $i$, the system stores all previous memory tokens in a memory cache, denoted as $M_i = [m_1, \ldots, m_i]$. We employ a self-retrieval mechanism to update the current memory token $m_i$. Specifically, we treat $m_i$ as a query and the concatenated memory cache $M_i$ as key and value. The model performs a standard multi-head cross-attention operation to integrate information from previous timesteps into the current memory state, yielding the updated memory token
$m_{i+1} = \text{Softmax}\left({W_i^Q m_i (W_i^K M_i)^\top}/{\sqrt{d_k}}\right) W_i^V M_i$, 
% \begin{equation}
% \small
%     \begin{aligned}
%     m_{i+1} = \text{Softmax}\left(\frac{W_i^Q m_i (W_i^K M_i)^\top}{\sqrt{d_k}}\right) W_i^V M_i,
%     \end{aligned}
% \end{equation}
where $W_i^Q, W_i^K, W_i^V$ are weight matrices for query, key and value, respectively.
% compute the query representation $Q_i = W^Q_i m_i$, where $W^Q_i$ is the query weight matrix at timestep $i$. Similarly, we construct the key $K_i$ and value $V_i$ representations by applying the key and value weight matrices $W^K_i$ and $W^V_i$ respectively to , resulting in $K_i = W^K_i C_i$ and $V_i = W^V_i C_i$. Subsequently, the model performs a standard multi-head cross-attention operation to integrate information from previous timesteps into the current memory state, yielding the updated memory token $m_{i+1} = \text{Softmax}\left(Q_i K_i^\top/\sqrt{d_k}\right) V_i$.

\paragraph{Computational Complexity}
% In terms of computational complexity, 
For bridge layers, we consider three main components for the theoretical complexity: (i) the self-attention within each segment, which scales as $\mathcal{O}((C+M)^2)$, where $C$ is the segment length and $M$ is the length of memory tokens; (ii) the memory retrieval, which scales as $\mathcal{O}(MK)$; and (iii) the recurrent processing. Consequently, the overall time complexity of our approach is $\mathcal{O}(K^2)$, and the space complexity is $\mathcal{O}(K)$. For the LLM, the complexity is $\mathcal{O}(M^2)$.
In practice, the segment length $C$ is a constant that depends on the constraint of LLM. $K$ is one $M$-th of $L$, thus our segmentation approach effectively compresses semantic units to an extreme degree, thereby striking a favorable balance between computational efficiency and model efficacy. Moreover, The number of segments can be fixed to accommodate the constraints of the environment.

\section{Experiments}\label{sec:exp}

\subsection{Setup}
We utilize Vicuna-7B-v1.5
% \footnote{\url{https://huggingface.co/lmsys/vicuna-7b-v1.5}}
as the LLM and ViT-L/14 as the visual backbone following Video-LLaVA~\citep{videollava}. 
% using LanguageBind as visual encoders, which are initialized from ViT-L/14. 
Each frame is resized and cropped to a dimension of 224×224. The Memory Bridge Layers are based on a single-layer Transformer. Our model is trained and evaluated with 16 frames and 4 segments, following the same video data protocol as PLLaVA~\cite{pllava}.  \looseness=-1
% For the NIVAH test, we use the memory tokens as input to LLMs to evaluate their memory capabilities. 
% For additional implementation details, please refer to \textbf{Supplementary Materials}.
% \textbf{Appendix \ref{supp:implementation}}.

\subsection{Long-form Video Understanding}
\label{sec:longvid}

% \subsubsection{Setups}

% \paragraph{Inference Setups}
% During the inference, we set the input frames to 32, and set a dynamic segment strategy, which dynamically set the number of segments to the mulitplies to 8.

% \paragraph{Datasets}
% We have chosen to evaluate our approach using two of the most widely-used and recognized long-form video QA datasets. The first dataset, The second dataset,In addition,

\paragraph{Baselines}

We conduct a comparative analysis of two types of models: retrieval-based methods and generative video-language models, as elaborated in \Cref{sec:related}. To ensure fairness in our comparisons, we primarily focus on state-of-the-art models such as LLaVA-NeXT-Video-DPO~\citep{llavanext} and PLLaVA~\citep{pllava}, which utilize the same base model and video datasets as our approach. Other SoTA models, including MovieChat~\citep{moviechat}, MA-LMM~\citep{malmm}, VideoStreaming~\citep{streaminglm}, and Video-xl~\citep{videoxl}, are not consistently included in all benchmarks due to variations in training data, model configurations, and benchmark settings. 
% (refer to Supplementary Materials for further details)
Nevertheless, we evaluate the key compression settings of these baselines using the same data and model configurations as ours, as detailed in \Cref{sec:bench,sec:ablate}.
 
% \subsubsection{Results and Analysis}
% We have chosen to evaluate our approach using two of the most widely-used and recognized long-form video QA datasets.

% add data
\begin{table*}[ht!]
    \centering
    \begin{minipage}[t]{0.48\textwidth} % Adjust the minipage width to fit your content
        \resizebox{\textwidth}{!}{%
        \begin{tabular}{lccc}
        \toprule
            \textbf{Model} &  \textbf{LLM}  & \textbf{Frames} & \textbf{Accuracy} \\ 
            \midrule
            GPT4-o & OpenAI API & 16 & 72.2  \\ \midrule

            \multicolumn{3}{l}{\small{\textit{Retrieval-based Video-Language Models}}} \\
            
            LongViViT*~\citeyear{longvivit} & - & 256 & 56.8 \\ 
    
            MC-ViT-L*~\citeyear{mcvit} &  - & 128 & 62.5 \\ \midrule
    
            % \multicolumn{3}{l}{\small{\textit{Agent-based Models}}} \\
    
            % % ViperGPT &  GPT-3.5 &4 & 15.8 \\ 
    
            % LLoVi & GPT-4 & 180 & 57.6 \\
    
            % Video-Agent (Retrieval) & GPT-4 & 8.4 & 60.2 \\
    
            % Video-Agent (Generative) &  GPT-4 &900 & 61.6 \\ \midrule
    
            \multicolumn{3}{l}{\small{\textit{Generative Video-Language Models}}} \\
    
            SeViLA~\citeyear{sevila} & Flan-T5-XL & 32 & 25.8 \\
            mPLUG-Owl~\citeyear{mplug-owl} & LLaMA-7B & 5 & 33.8 \\
            Video-LLaVA~\citeyear{videollava} & Vicuna-7B & 8 & 40.2 \\ 
            LLaVA-NeXT-Video-DPO~\citeyear{llavanext} & Vicuna-7B & 32 & 41.6 \\
            PLLaVA~\citeyear{pllava} & Vicuna-7B & 16 (16) & 45.6 \\
            PLLaVA~\citeyear{pllava} & Vicuna-7B & 32 (16) & 43.8 \\
            % PLLaVA-13B & Vicuna-13B & 16 (16) & 54 \\
            % PLLaVA-34B & Yi-34B & 16 (16) & 61.6 \\
            % \textcolor{gray}{PLLaVA-13B} & \textcolor{gray}{Vicuna-13B} & \textcolor{gray}{16 (16)} & \textcolor{gray}{54} \\
            % \textcolor{gray}{PLLaVA-34B} & \textcolor{gray}{Yi-34B} & \textcolor{gray}{16 (16)} & \textcolor{gray}{61.6} \\

            \rowcolor{LightCyan} \textbf{\model(Ours)} & Vicuna-7B  & 32 (8)  & \textbf{53.8}  \\
            \bottomrule
        \end{tabular}
        }
        \caption{\textbf{Results on Subset of EgoSchema under zero-shot setting.}  $^*$: the model has been fine-tuned using the training data from EgoSchema. $(n)$: $n$ frames are used in training.}
        \label{tab:egoschema}
    \end{minipage}%
            % \vspace{-.1in}
    \hfill % this will insert a non-breakable space between minipages and push them to the sides of text area
    \begin{minipage}[t]{0.48\textwidth} % Adjust the minipage width to fit your content
        \resizebox{\textwidth}{!}{%
        \begin{tabular}{lcccc}
        \toprule
            \textbf{Model} & \textbf{Temporal} & \textbf{Causal} & \textbf{Description} & \textbf{All} \\ \midrule

            GPT4-o & 70.3 & 78.0 & 80.8 & 76.0 \\ \midrule
            
            \multicolumn{5}{l}{\small{\textit{Retrieval-based Video-Language Models}}} \\
            \midrule
            
            AIO*~\citeyear{aio} & 48.0 & 48.6 & 63.2 & 50.6 \\ 

            VQA-T*~\citeyear{vqat} & 49.6 & 51.5 & 63.2 & 52.3 \\
    
            ATP*~\citeyear{atp} & 50.2 & 53.1 & 66.8 & 54.3 \\ 
    
            VGT*~\citeyear{vgt} & 52.3 & 55.1 & 64.1 & 55.0 \\
            
            MIST-CLIP*~\citeyear{mist} & 56.6 & 54.6 & 66.9 & 57.1 \\ 
            % InternVideo & 48.0 & 43.4 & 65.1 & 49.1 \\ 
            
            \midrule
    
            % \multicolumn{5}{l}{\small{\textit{Agent-based Models}}} \\
    
            % AssistGPT & 51.4 & 60.0 & 67.3 & 58.4 \\
            % LLoVi & 61.0 & 69.5 & 75.6 & 67.7 \\
            % VideoAgent (Retrieval) & 64.5 & 72.7 & 81.1 & 71.3 \\ \midrule
    
            \multicolumn{5}{l}{\small{\textit{Generative Video-Language Models}}} \\
            % \multicolumn{5}{l}{\small{\textit{Finetuned}}} \\
    
            % BLIP2* & 64.9 & 69.7 & 79.4 & 69.6 \\
            % SEVILA* & 69.4 & 74.2 & 81.3 & 73.8 \\
            % PLLaVA* & 62.2 & 68.5 & 79.7 & 68.2 \\ 
            % \model(Ours)*  & 66.8 & 71.6 & 78.4 & 71.1 \\ \midrule
            % % \multicolumn{5}{l}{\tiny{\small{Zero-shot}}} \\

            SeViLA~\citeyear{sevila} & 61.5 & 61.3 & 75.6 & 63.6 \\
            LLaMA-VID~\citeyear{llamavid} & 53.8 & 60.0 & 73.0 & 59.5 \\
            Video-LLaVA~\citeyear{videollava} & 56.9 & 61.0 & 75.0 & 61.3 \\
            
            LLaVA-NeXT-Video-DPO~\citeyear{llavanext} & 55.6 & 61.0 & 73.9 & 61.3 \\
            PLLaVA*~\citeyear{pllava} & 62.2 & 68.5 & \textbf{79.7} & 68.2 \\
            
             % \model(Ours)  & 57.44 & 60.30 & 73.87 & 61.48 \\
             \rowcolor{LightCyan} \textbf{\model(Ours)*}  & \textbf{66.8} & \textbf{71.6} & 78.4& \textbf{71.1} \\
             \bottomrule
        \end{tabular}
        }
        \caption{\textbf{Comparison accuracy on NExT-QA.} $^*$ indicates that the instruction data includes the training data from NExTQA.}
        \label{tab:nextqa}
    \end{minipage}
                % \vspace{-.1in}

\end{table*}

\begin{table*}[t!]
        \resizebox{\textwidth}{!}{
        \begin{tabular}{l cc ccccc ccccc ccccc cccccc }
            \toprule
            \textbf{Method} & \makecell{\textbf{Vision}\\\textbf{Encoder}} & \makecell{\textbf{LLM}\\\textbf{Size}} 
            & \textbf{AS} & \textbf{AP} & \textbf{AA} & \textbf{FA} & \textbf{UA} 
            & \textbf{OE} & \textbf{OI} & \textbf{OS} & \textbf{MD} & \textbf{AL} 
            & \textbf{ST} & \textbf{AC} & \textbf{MC} & \textbf{MA} & \textbf{SC} 
            & \textbf{FP}  & \textbf{CO} & \textbf{EN} & \textbf{ER} & \textbf{CI} & \textbf{Avg.}\\
            \midrule
            GPT-4V & GPT-4V & / & 55.5 & 63.5 & 72.0 & 46.5 & 73.5 & 18.5 & 59.0 & 29.5 & 12.0 & 40.5 & 83.5 & 39.0 & 12.0 & 22.5 & 45.0 & 47.5 & 52.0 & 31.0 & 59.0 & 11.0 & 43.5\\
            \midrule

             mPLUG-Owl-I~\citeyear{mplug-owl} & ViT-L & 7B  & 25.0 & 20.0 & 44.5 & 27.0 & 23.5 & 36.0 & 24.0 & 34.0 & 23.0 & 24.0 & 34.5 & 34.5 & 22.0 & 31.5 & 40.0 & 24.0 & 37.0 & 25.5 & 21.0 & 37.0& 29.4 \\
            LLaMA-Adapter~\citeyear{llamaadapter} & ViT-B & 7B & 23.0 & 28.0 & 51.0 & 30.0 & 33.0 & 53.5 & 32.5 & 33.5 & 25.5 & 21.5 & 30.5 & 29.0 & 22.5 & 41.5 & 39.5 & 25.0 & 31.5 & 22.5 & 28.0 & 32.0 & 31.7 \\
            BLIP2~\citeyear{blip2} & ViT-G & 2.7B & 24.5 & 29.0 & 33.5 & 17.0 & 42.0 & 51.5 & 26.0 & 31.0 & 25.5 & 26.0 & 32.5 & 25.5 & 30.0 & 40.0 & 42.0 & 27.0 & 30.0 & 26.0 & 37.0 & 31.0 & 31.4 \\
            Otter-I~\citeyear{otter} & ViT-L & 7B & 34.5 & 32.0 & 39.5 & 30.5 & 38.5 & 48.5 & 44.0 & 29.5 & 19.0 & 25.5 & 55.0 & 20.0 & 32.5 & 28.5 & 39.0 & 28.0 & 27.0 & 32.0 & 29.0 & 36.5& 33.5  \\
            MiniGPT-4~\citeyear{minigpt4} & ViT-G & 7B & 16.0 & 18.0 & 26.0 & 21.5 & 16.0 & 29.5 & 25.5 & 13.0 & 11.5 & 12.0 & 9.5 & 32.5 & 15.5 & 8.0 & 34.0 & 26.0 & 29.5 & 19.0 & 9.9 & 3.0& 18.8  \\
            InstructBLIP~\citeyear{instructblip} & ViT-G & 7B & 20.0 & 16.5 & 46.0 & 24.5 & 46.0 & 51.0 & 26.0 & 37.5 & 22.0 & 23.0 & 46.5 & \textbf{42.5} & 26.5 & 40.5 & 32.0 & 25.5 & 30.0 & 25.5 & 30.5 & 38.0 & 32.5 \\
            LLaVA~\citeyear{llava} & ViT-L & 7B  & 28.0 & 39.5 & 63.0 & 30.5 & 39.0 & 53.0 & 41.0 & 41.5 & 23.0 & 20.5 & 45.0 & 34.0 & 20.5 & 38.5 & 47.0 & 25.0 & 36.0 & 27.0 & 26.5 & 42.0& 36.0 \\

            \midrule
            
            Video-LLaMA~\citeyear{videollama} & CLIP-G & 7B &  27.5 &25.5 & 51.0 & 29.0 & 39.0 & 48.0 & 40.5 & 38.0 & 22.5 & 22.5 & 43.0 & 34.0 & 22.5 & 32.5 & \hlttrd{45.5} & 32.5 & 40.0 & 30.0 & 21.0 & 37.0 & 34.1 \\
            LLaMA-Adapter~\citeyear{llamaadapter} & ViT-B & 7B & 23.0 & 28.0 & 51.0 & 30.0 & 33.0 & 53.5 & 32.5 & 33.5 & 25.5 & 21.5 & 30.5 & 29.0 & 22.5 & 41.5 & 39.5 & 25.0 & 31.5 & 22.5 & 28.0 & 32.0 & 31.7 \\
            Video-ChatGPT~\citeyear{videochatgpt} & ViT-L & 7B & 23.5 & 26.0&  62.0 & 22.5 & 26.5 & 54.0 & 28.0 & \hltse{40.0} & 23.0 & 20.0 & 31.0 & 30.5 & 25.5 & 39.5 & \hlt{48.5} & 29.0 & 33.0 & 29.5 & 26.0 & 35.5 & 32.7 \\
            
            VideoChat~\citeyear{videochat} & CLIP-G & 7B & 33.5 & 26.5 & 56.0 & 33.5 & 40.5 & 53.0 & 40.5 & 30.0 & \hlttrd{25.5} & \hltse{27.0} & 48.5 & 35.0 & 20.5 & 42.5 & \hltse{46.0} & 26.5 & 41.0 & 23.5 & 23.5 & 36.0 & 35.5 \\
            % VideoChat2 & UMT-L & 7B & 66.0 & 47.5 & \textbf{83.5} & \textbf{49.5} & 60.0 & 58.0 & \textbf{71.5} & \textbf{42.5} & 23.0 & 23.0 & 88.5 & 39.0 & 42.0 & 58.5 & 44.0 & 49.0 & 36.5 & 35.0 & 40.5 & \textbf{65.5} & 51.1 \\
            VideoChat2$^\beta$~\citeyear{mvbench} & UMT-L & 7B & \hlt{66.0} & \hlttrd{47.5} & \hlttrd{83.5} & \hlt{49.5} & \hltse{60.0} & \hlttrd{58.0} & \hlt{71.5} & \hlt{42.5} & 23.0 & 23.0 & \hlt{88.5} & 39.0 & \hlttrd{42.0} & \hlttrd{58.5} & 44.0 & \hlt{49.0} & 36.5 & \hlt{35.0} &\hlttrd{40.5} & \hlt{65.5} & \hltse{51.1} \\

            % ST-LLM & BLIP2 & 7B  & 66.0 & 53.5 & 84.0 & 44.0 & 58.5 & 80.5 & 73.5 & 38.5 & 42.5 & 31.0 & 86.5 & 36.5 & 56.5 & 78.5 & 43.0 & 44.5 & 46.5 & 34.5 & 41.5 & 58.5 & 54.9 \\

            PLLaVA 7B$^\alpha$~\citeyear{pllava}  & ViT-L & 7B  & \hltse{58.0} & \hltse{49.0} & 55.5 & 41.0 & \hlt{61.0} & 56.0 & \hltse{61.0} & 36.0 & 23.5 & \hlttrd{26.0} & \hlttrd{82.0} & \hlttrd{39.5} & \hlttrd{42.0} & 52.0 & 45.0 & \hltse{42.0} & \hlt{53.5} & \hltse{30.5} & \hlt{48.0} & 31.0 & 46.6 \\
            \textcolor{gray}{PLLaVA 13B}$^\alpha$~\citeyear{pllava} & \textcolor{gray}{ViT-L} & \textcolor{gray}{13B} & \textcolor{gray}{66.0} & \textcolor{gray}{53.0} & \textcolor{gray}{65.5} & \textcolor{gray}{45.0} & \textcolor{gray}{65.0} & \textcolor{gray}{58.0} & \textcolor{gray}{64.5} & \textcolor{gray}{35.5} & \textcolor{gray}{23.5} & \textcolor{gray}{30.0} & \textcolor{gray}{85.0} & \textcolor{gray}{39.5} & \textcolor{gray}{45.5} & \textcolor{gray}{57.0} & \textcolor{gray}{47.5} & \textcolor{gray}{49.5} & \textcolor{gray}{49.0} & \textcolor{gray}{33.0} & \textcolor{gray}{53.0} & \textcolor{gray}{37.0} & \textcolor{gray}{50.1} \\
            % \textcolor{gray}{PLLaVA 34B} & \textcolor{gray}{ViT-L} & \textcolor{gray}{34B} & \textcolor{gray}{67.5} & \textcolor{gray}{53.0} & \textcolor{gray}{82.0} & \textcolor{gray}{47.0} & \textcolor{gray}{79.0} & \textcolor{gray}{68.5} & \textcolor{gray}{67.5} & \textcolor{gray}{36.5} & \textcolor{gray}{37.5} & \textcolor{gray}{49.5} & \textcolor{gray}{91.0} & \textcolor{gray}{40.5} & \textcolor{gray}{43.0} & \textcolor{gray}{70.0} & \textcolor{gray}{51.5} & \textcolor{gray}{50.0} & \textcolor{gray}{66.5} & \textcolor{gray}{39.5} & \textcolor{gray}{63.5} & \textcolor{gray}{59.0} & \textcolor{gray}{58.1} \\
    
            % PLLaVA 13B & ViT-L & 13B & 66.0 & 53.0 & 65.5 & 45.0 & 65.0 & 58.0 & 64.5 & 35.5 & 23.5 & 30.0 & 85.0 & 39.5 & 45.5 & 57.0 & 47.5 & 49.5 & 49.0 & 33.0 & 53.0 & 37.0 & 50.1 \\
            % PLLaVA 34B & ViT-L & 34B & 67.5 & 53.0 & 82.0 & 47.0 & 79.0 & 68.5 & 67.5 & 36.5 & 37.5 & 49.5 & 91.0 & 40.5 & 43.0 & 70.0 & 51.5 & 50.0 & 66.5 & 39.5 & 63.5 & 59.0 & 58.1 \\
    
            \textbf{\model$^\alpha$ (Ours)}  & ViT-L & 7B  & 52.0 &\hlt{50.5} & \hltse{85.5} & \hlttrd{42.5} & 51.0 & \hltse{69.5} & 56.0 & \hlttrd{38.5} & \hltse{41.0} & 24.0 & 69.5 & \hltse{40.0} & \hltse{48.0} & \hltse{71.5} & 43.5 & 34.5 & \hlttrd{41.5} & 29.5 & 38.0 & \hltse{60.0} & \hlttrd{49.33} \\

            \textbf{\model$^\beta$ (Ours)}  & ViT-L & 7B  & \hlttrd{54.5} & 47.0 & \hlt{86.5} & \hltse{44.5} & \hlttrd{52.0} & \hlt{79.0} & \hlttrd{58.5} & 32.0 & \hlt{47.0} & \hlt{33.0} & \hltse{82.5} & \hlt{40.5} & \hlt{52.0} & \hlt{82.0} & 40.5 & \hlttrd{37.5} & \hltse{43.0} & \hlttrd{31.0} & \hltse{42.5} & \hltse{60.0} & \hlt{52.5} \\
            \bottomrule
        \end{tabular}}
        \caption{\textbf{Results on MVBench~\citep{mvbench} multi-choice question answering.} 
        % We list GPT-4V in the first row group as a reference. The second row group includes image-based MLLMs. The third row group includes video-based MLLMs. 
        We highlight top-3 results among all 7B models of each category in \hlt{purple}. $\alpha$: training data from \citet{pllava}. $\beta$: training with data from \citet{mvbench}.}
        \label{tab:mvbench}
        \vspace{-.1in}
\end{table*}

\begin{table}[t!]
    \centering
    % \vspace{-1.0em}
    \resizebox{\linewidth}{!}{%
    \begin{tabular}{lccccccc}
    \toprule
        
        \textbf{Model} & \textbf{Data}    & \textbf{Short} & \textbf{Medium} & \textbf{Long} & \textbf{All.} & \textbf{Comp.} \\ \midrule

        % VideoLLaVA~\citeyear{videollava} & VideoChatGPT & Vicuna-7B & 8 & 45.3 & 38.0 & 36.2 & 39.9 \\ 
        % VideoLLaVA~\citeyear{videollava} & VideoChatGPT & Vicuna-7B & 8 & 41.0 & 35.78 & 36.44 & 37.74 \\ 
        LLaVA-NeXT-Vicuna~\citeyear{llavanext} & LLaVA-NeXT  & 35.26 & 37.44 & 32.88 & 35.44 & 1.00 \\
        % \textcolor{gray}{LLaVA-NeXT-Qwen2~\citeyear{longva}} & \textcolor{gray}{LLaVA-NeXT}  & \textcolor{gray}{Qwen2-7B} & \textcolor{gray}{58.0} & \textcolor{gray}{47.0} & \textcolor{gray}{43.4} & \textcolor{gray}{49.5} \\
        % \textcolor{gray}{LongVA~\citeyear{longva}} & \textcolor{gray}{LLaVA-NeXT}  & \textcolor{gray}{Qwen2-7B} & \textcolor{gray}{61.1} & \textcolor{gray}{50.4} & \textcolor{gray}{46.2} & \textcolor{gray}{52.6} \\
        VideoChat2~\citeyear{videochat} & VideoChat2  & - & - & - & 33.7 & 0.49 \\
        % \textcolor{gray}{VideoChat2-Mistral~\citeyear{videochat}} & \textcolor{gray}{VideoChat2} & \textcolor{gray}{Mistral-7B} & \textcolor{gray}{16} & \textcolor{gray}{48.3} & \textcolor{gray}{37.0} & \textcolor{gray}{33.2} & \textcolor{gray}{39.5} \\
        PLLaVA~\citeyear{pllava} & PLLaVA   & 46.44 & 38.00 & 33.22 & 38.22 & 0.25 \\ 
        % PLLaVA~\citeyear{pllava} & PLLaVA & Vicuna-7B  & 48.44 & 39.55 & 36.22 & 41.41 \\
        % 38.22 46.44 38 33.22
        \textbf{\model $\alpha$ (Ours)} & PLLaVA   & 46.11 & 38.44 & 34.22 & 39.59  & 0.06  \\
        \textbf{\model $\beta$ (Ours)} & VideoChat2  & \hlt{49.22} & \hlt{39.11} & \hlt{35.89} & \hlt{41.41} & 0.06  \\
         \bottomrule
         % \multicolumn{5}
    \end{tabular}
    }
    % \vspace{-1.0em}
    \caption{\textbf{Results on VideoMME}. We list models that use the same training data and Vicuna-7B backbones for fair comparison, for existing SoTA models like Video-XL~\citep{videoxl}, VideoStreaming~\citep{streaminglm} are trained on self-constructed dataset. Comp: Compression.}
    \label{tab:videomme}
    \vspace{-.2in}
\end{table}

\paragraph{Results on EgoSchema} EgoSchema~\citep{egoschema} consists of egocentric videos, each averaging \textbf{180 seconds} in length. This video QA dataset focuses on aspects such as understanding, reasoning, and long-term memory. In our experiment, we follow the precedent set by previous studies and use the public subset for evaluation. The results are presented in \Cref{tab:egoschema}. Overall, our method significantly outperforms current generative video language models trained on similar data, demonstrating robust performance compared to other approaches and confirming its efficacy. Specifically, we compare our method with PLLaVA~\cite{pllava}, which shares the same training data, LLM backbones, and input number of frames. 
Our method shows significant improvements over PLLaVA, indicating its superiority in understanding long egocentric videos. While our method does not yet match the performance of fine-tuned retrieval-based methods, we plan to apply our approach to larger language models to bridge this performance gap.

% While our method does not yet match the performance of agent-based methods that integrate a collection of specialized visual
% experts from various training domains and powerful language models like GPT-4, which could lead to heavy inference time, it shows promising potential to achieve similar results more efficiently and effectively. As for retrieval-basd methods, most of which would  Looking ahead, we plan to apply our method to larger language models in an effort to bridge this performance gap.

\paragraph{Length Extrapolation}

% \begin{wrapfigure}{r}{.5\linewidth}
\begin{figure}[t!]
    \centering
    % \vspace{-2em}
    \includegraphics[width=\linewidth]{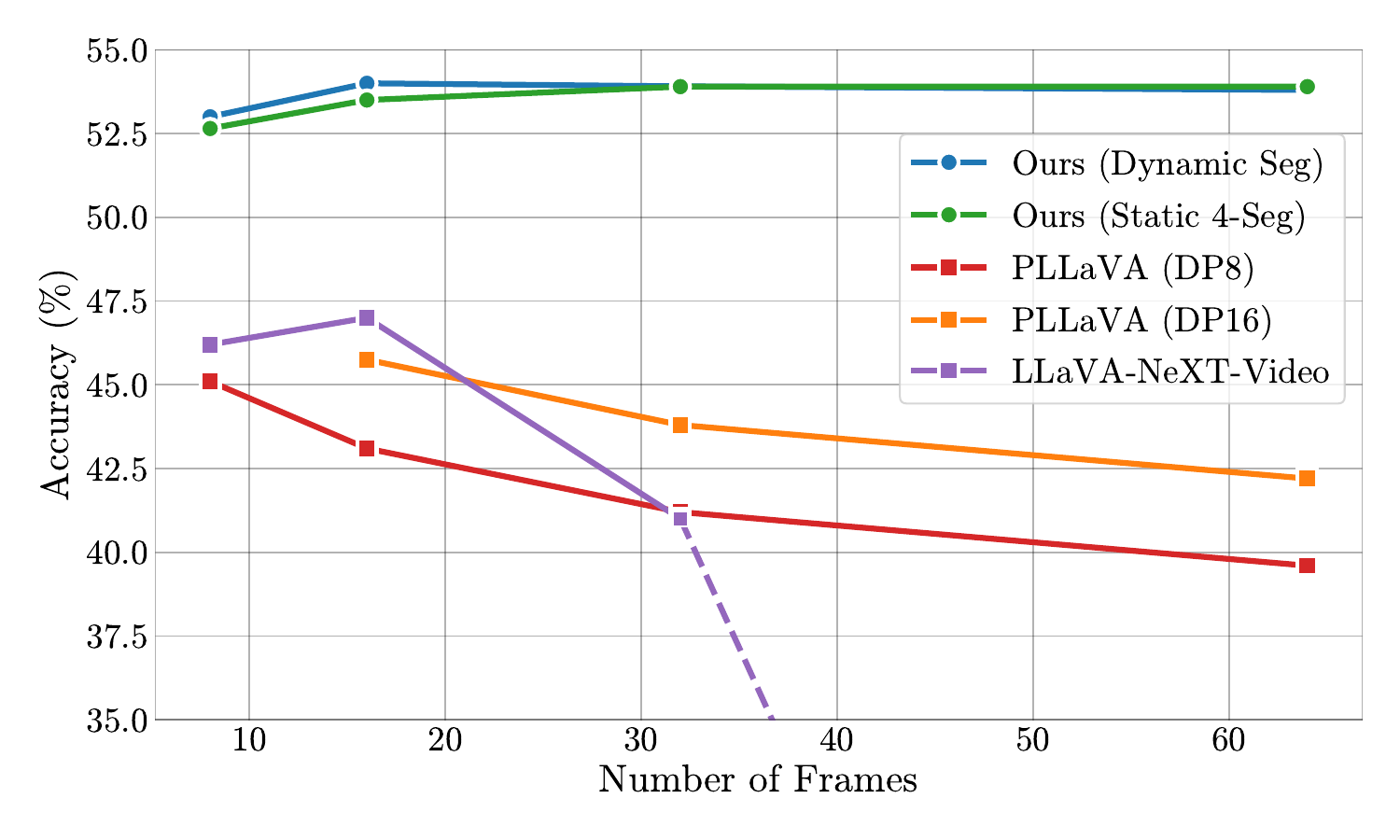}
    \vspace{-2.5em}
    \caption{\textbf{Length extrapolation results} on EgoSchema dataset.}
    \label{fig:length}
    \vspace{-.1in}
% \end{wrapfigure}
\end{figure}
The model is trained on 16-frame sequences, divided into 4 segments. However, in real-world scenarios, videos can be significantly longer than this training configuration. To demonstrate \model's ability to extrapolate to longer videos, we conducted experiments on EgoSchema under two conditions: 1) dynamic segments, which adaptively control the number of segments based on the SceneTiling threshold, and 2) static segments, fixed at 4 segments.  Results in \Cref{fig:length} reveal that dynamic segments are more effective than static segments, especially for shorter videos, indicating that our method can effectively maintain an appropriate number of segments. However, as video length increases, the performance of dynamic segments declines, notably at the 32-frame mark, where both strategies use four segments. Beyond this point, increasing the number of segments results in diminishing returns, likely due to the domain gap from training on shorter videos. To address this issue, we plan to fine-tune our models on longer videos for more substantial improvements. Overall, compared to PLLaVA, our method maintains consistent performance as the input length increases. In summary, our approach effectively extracts key information from videos, outperforming the simple pooling strategies used for memory consolidation in existing methods.

% \begin{figure}[ht!]
%     \centering
%     \small
%     % \begin{minipage}{.47\linewidth}
%     % % \vspace{-.2in}
%     % \includegraphics[width=\linewidth]{figures/memory.pdf}
%     % \caption{\textbf{GPU Memory Cost.} We apply all the experiments on a single NVIDIA A800 GPU.}
%     % \label{fig:memory}
%     % \vspace{-0.1in}
%     % \end{minipage}\hfill
%     \begin{minipage}{.47\linewidth}
%             \centering
%     \includegraphics[width=\linewidth]{figures/num_frame.pdf}
%     \caption{\textbf{Length extrapolation results} on EgoSchema dataset.}
%     \label{fig:length}
%         \vspace{-0.1in}
%     \end{minipage}
% \end{figure}

\paragraph{Results on NExTQA}  NExTQA \citep{nextqa}, featuring daily-life videos that average 45 seconds in length, is designed to test a variety of question types, specifically temporal, causal, and descriptive questions. We applied our method to NExTQA to evaluate its temporal grounding ability. To maintain consistency with established benchmarks, we used the validation set for evaluation. In \Cref{tab:nextqa}, we present the comprehensive results of our analysis. For a fair comparison, our primary benchmark is against PLLaVA, which includes instruction data from the NExTQA training set. Our method surpasses PLLaVA by $2.9$ points. Notably, our approach demonstrates a significant enhancement in the temporal setting, achieving a $4.6$ point improvement over PLLaVA. These results indicate that our scene-segment aware method effectively improves the model's temporal grounding ability by compressing abundant information within scenes that share high semantic similarity.
% Overall, our method exhibits robust performance in this setting compared to all other methods tested. This underscores the benefits of our method's enhanced long-term memory retention.

% \subsubsection{MovieChat}
% % moviechat
% \begin{table}[ht!]
%     \centering
%     % \vspace{-1.0em}
%     \resizebox{\linewidth}{!}{%
%     \begin{tabular}{lccccc}
%     \toprule
        
%         \multirow{2}{*}{\textbf{Methods}} & \multicolumn{2}{c}{\textbf{Global}} & \multicolumn{2}{c}{\textbf{Breakpoint}} &  \\ 
%         \cmidrule{2-3} \cmidrule{4-5} &  Accuracy & Score & Accuracy & Score \\ \midrule

%         VideoChat & 57.8 & 3.00 & 46.1 & 2.29 \\
%         Video-LLaMA & 51.7 & 2.67 & 39.1 & 2.04 \\
%         VideoChatGPT & 47.6 & 2.55 & 48.0 & 2.45 \\
%         MovieChat & 62.3 & 3.23 & 48.3 & 2.57 \\
%         % MovieChat+ & 71.2 & 3.51 & 49.6 & 2.62 \\
%         LLaVA-NeXT-Video-DPO & 63.6 & 3.38 & 30.5 & 1.60 \\
%         PLLaVA & 64.0 & 3.2 & 37.2 & 2.03 \\
%         \model(Ours)  & - & - & - & - \\ 
%         \bottomrule
%          % \multicolumn{5}
%     \end{tabular}
%     }
%     % \vspace{-1.0em}
%     \caption{\textbf{Comparison accuracy of long-form video QA on MovieChat}}
%     \label{tab:moviechat}
%     % \vspace{-.2in}
% \end{table}

\paragraph{Results on Long Context VideoQA} We evaluate our methods on the long video benchmark, VideoMME~\citep{videomme}, which contain videos ranging from 11 seconds to \textbf{1 hour} in length. 
The results are illustrated in \Cref{tab:videomme}. We compare methods using the same training data and LLM backbones. Compared to PLLaVA-B~\cite{pllava} and VideoChat-2~\cite{videochat}, our method shows improvements on both benchmarks. Notably, our method demonstrates consistent improvements on long videos in Video-MME with much less compression rate.

\subsection{Comprehensive Video Understanding}
% \paragraph{Results on Comprehensive Video Understanding Benchmark}
We evaluated our method using the comprehensive video understanding benchmark MVBench~\citep{mvbench}. As shown in \Cref{tab:mvbench}, our approach maintains strong performance across general video understanding tasks. Remarkably, with the same training data as PLLaVA, our method achieves performance comparable to a 13B-level model. Our method effectively extracts information from both short and long videos. To assess scalability, we trained our model on the VideoChat2~\citep{videochat} dataset. The results, presented at the bottom of \Cref{tab:mvbench}, demonstrate that our model, when trained on larger video datasets, improves accuracy on MVBench by 3.17 points and outperforms VideoChat2, which was trained on the same dataset.

\subsection{Planning Tasks}
\label{sec:egoplan}
% egoplan

% \subsubsection{Setups}

% \paragraph{Inference Setups} For the planning tasks, we have configured the input to consist of 16 frames. Additionally, we have adopted the same dynamic segmentation strategy that is utilized in video QA.

% \paragraph{Datasets} 

\paragraph{Baselines} 
% Given the relatively brief duration of the input videos of the current planning benchmark, our comparative analysis includes both image-language and video-language models. 
The original protocol dictated the selection of a single frame corresponding to each action. To refine this approach and enhance the evaluation process, we introduce a smoother method. This involves segmenting the entire video into intervals based on predefined timesteps. This revised method is applied in the evaluation of PLLaVA, LLaVA-NeXT-Video-DPO, and \model. 
% Detailed descriptions for each baseline model are discussed in \Cref{app:baseline}.

% \subsubsection{Results and Analysis}
% \begin{table}[ht!]
    % \vspace{-.2in}
% \end{table}

\paragraph{Results on EgoPlan~\citep{egoplan}} The EgoPlan dataset \citep{egoplan} was developed as an egocentric question-answering benchmark tailored for embodied planning tasks, comprising 3,355 questions. The evaluation follows the framework established in the original study, utilizing the probability \( p(a|v,l) \) to identify the most suitable answer candidates. In \Cref{tab:egoplan}, we demonstrate that our model surpasses all other video-language models in performance. This suggests that our model's use of memory significantly enhances its planning capabilities compared to methods focused on the current stage. 
% We additionally show the qualitative results in \Cref{fig:case-egoplan}.
% While our approach does not outperform certain image-language models, we attribute this to the constraints of the current benchmark, which features brief action sequences and carefully curated frame-action pairs. Our goal is to develop more challenging benchmarks for egocentric planning and to adapt our method for real-time planning tasks. 
We are confident that our method holds great promise for generalizing to practical, real-world scenarios.

% \paragraph{Qualatitive Results}

\begin{table}[]
    \centering
    \resizebox{\linewidth}{!}{%
    \begin{tabular}{lcc}
    \toprule

        \textbf{Model} & \textbf{LLM} & \textbf{Accuracy} \\ \midrule
        GPT-4V & OpenAI API & 37.98 \\ \midrule
        
        % \multicolumn{3}{l}{\small{\textit{Image-Language Model}}} \\
        % Qwen-VL-Chat~\citep{qwen} & Qwen-7B & 26.32 \\
        % LLaVA-1.5~\citep{llava15} & Vicuna-7B & 26.80 \\
        % SEED-LLaMA~\citep{seedllama} & LLaMA2-Chat-13B & 29.93 \\ 
        % InternLM-Xcomposer~\citep{internlm} & InternLM-7B & 34.4 \\ \midrule
        % \multicolumn{3}{l}{\small{\textit{Video-Language Model}}} \\
        % VideoChatGPT~\citep{videochatgpt} & LLaMA-7B & 26.35 \\ 

        % Valley~\citep{vally} & LLaMA-13B & 26.17 \\

        VideoLLaMA~\citep{videollama} & LLaMA2-Chat-7B & 29.85 \\
        LLaVA-NeXT-Video~\citep{llavanext} & Vicuna-7B & 28.96 \\
        PLLaVA~\citep{pllava} & Vicuna-7B & 30.26 \\

    \rowcolor{LightCyan}\textbf{\model(Ours)}  & Vicuna-7B & \textbf{32.32}  \\
         \bottomrule
         % \multicolumn{5}
    \end{tabular}
    }
    % \vspace{-1.0em}
    \captionof{table}{\small \textbf{Results on EgoPlan under Zero-shot setting.}}
    \label{tab:egoplan}
    \vspace{-.1in}
\end{table}

\subsection{Streaming Caption}
% \label{supp:streaming}
Streaming dense video captions \citep{video-online,streamingcap} involves generating captions for videos in real-time, without the need to process the entire video sequence beforehand. The primary challenge in this task is determining the exact timestamps to predict event captions. Most existing methods rely on special tokens, annotated as the end of an action, for training. Our approach introduces the SceneTiling algorithm, which can automatically identify the break points in a streaming video and generate captions without requiring special training tokens. To enhance the efficiency of our method, we calculate the depth score using only the left similarity: \( d_i = \left(cl_i - c_i\right) / 2 \). 
% In \Cref{fig:streaming}, we present qualitative results of our method applied to a streaming video. 
This demonstrates that our method can effectively detect scene changes and automatically generate event captions.

% needle-in-haystack
\subsection{Stress Test: ``Needle In a Video Haystack''}
\label{sec:bench}

% \begin{figure*} [thbp!]
%     \centering
%     \begin{tabular}{cc}
%          \textbf{Needle:}  \textit{A young man is sitting on a piece of cloud in the sky, reading a book.} \\
%              \includegraphics[width=\textwidth]{figures/case_needle.pdf}
%     \end{tabular}
%     % \includegraphics[width=\textwidth]{figures/case_needle.pdf}
%     \caption{\textbf{Example of \benchshort.} For the text needle, the description is appended directly to the video; for the image and video needles, the corresponding image and video clips are inserted into the video haystack.}
%     \label{fig:neeld}
% \end{figure*}

\begin{figure*}[t!]
\small
    \centering
    % \vspace{-2em}
    \newlength{\imagewidth}
    \setlength{\imagewidth}{.49\linewidth}
    % \resizebox{\linewidth}{!}{
    \begin{tabular}{cc}
        %     \scriptsize PLLaVA~\citep{pllava} & \scriptsize LLaVA-NeXT-Video-DPO~\citep{llavanext} \\ 
        % \includegraphics[width=\imagewidth]{figures/pllava_new.pdf} & \includegraphics[width=\imagewidth]{figures/llavanext_new.pdf} \\ 
                \scriptsize MA-LMM~\citep{malmm} & \scriptsize LongVA~\citep{longva} \\
        \includegraphics[width=\imagewidth]{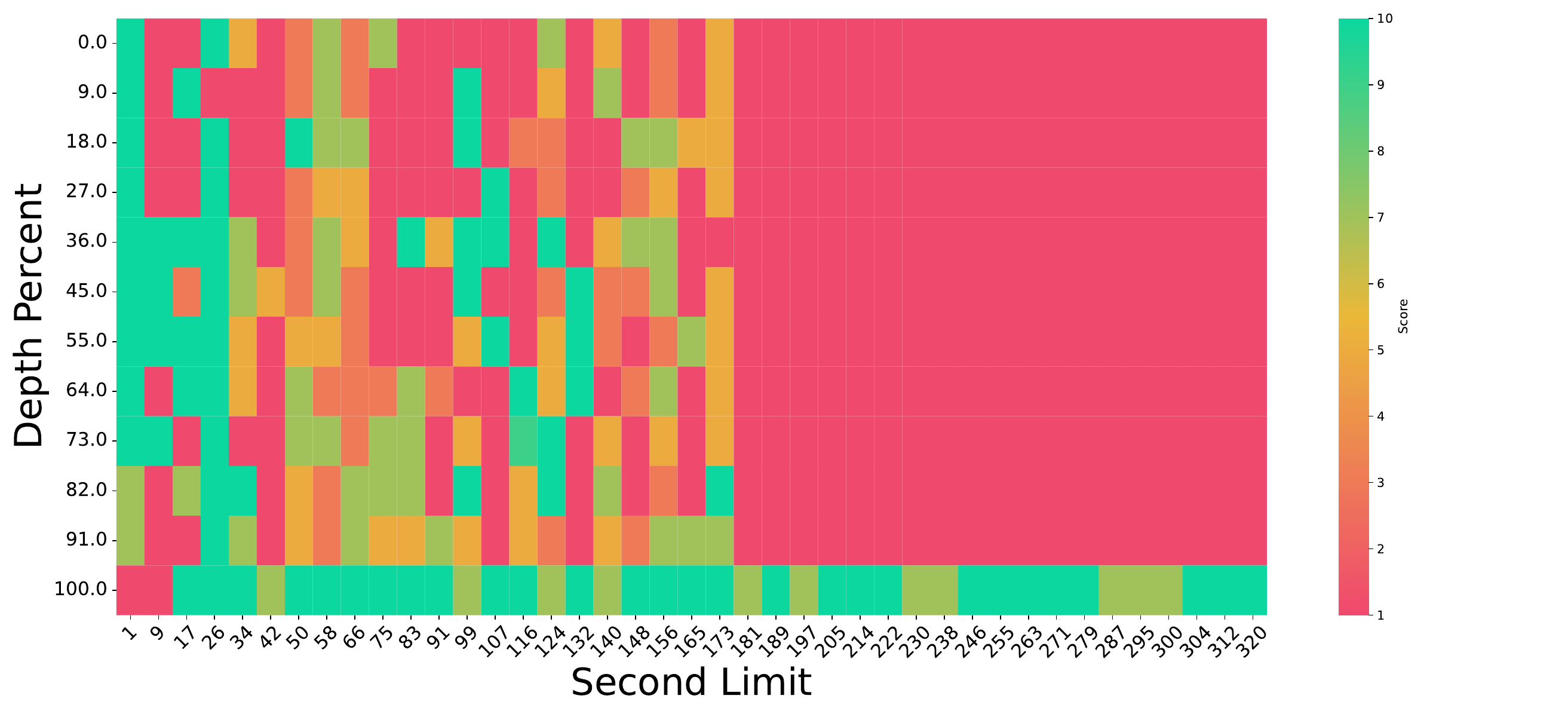} & \includegraphics[width=\imagewidth]{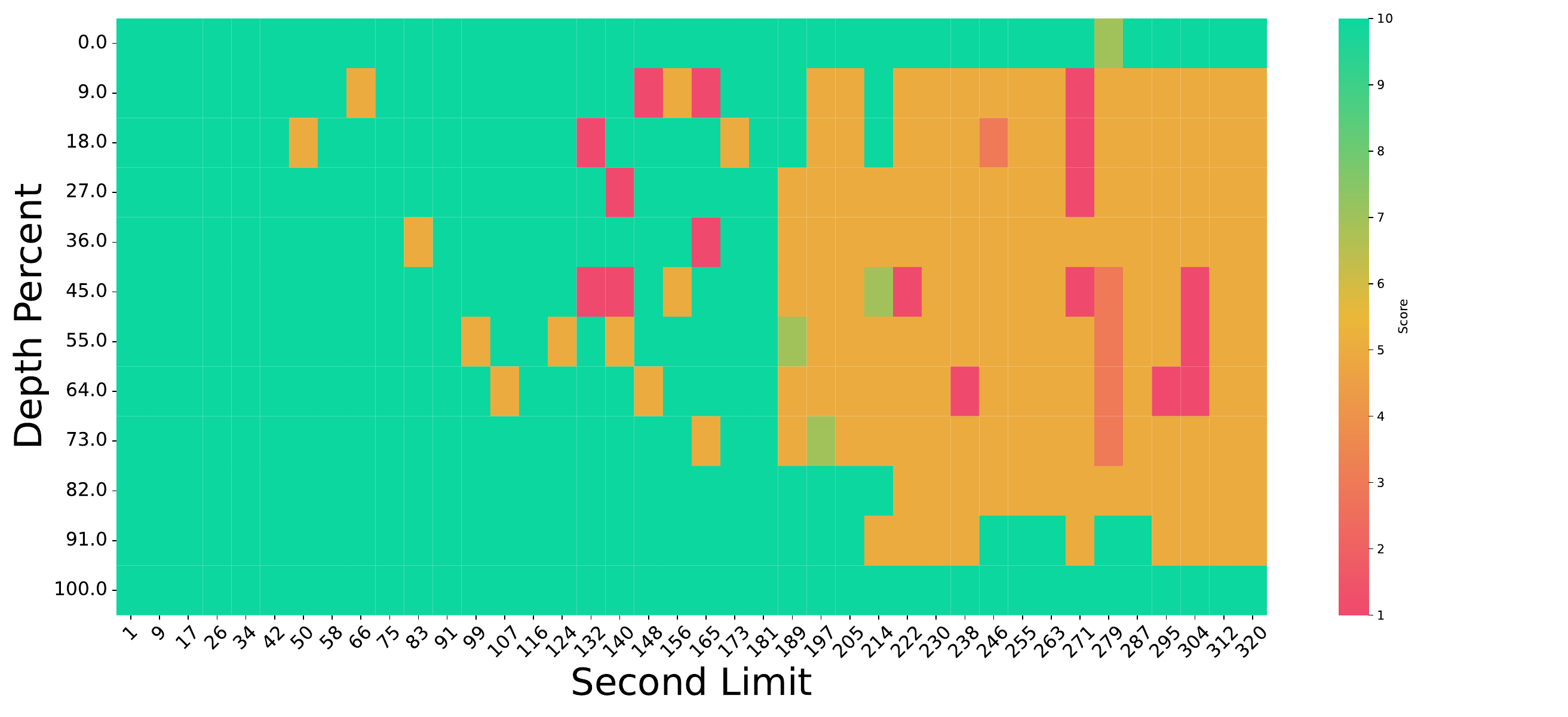} \\ 
                \scriptsize \textbf{\model-Mem} w.o. retrieval &\scriptsize \textbf{\model-Mem}-Full \\
        \includegraphics[width=\imagewidth]{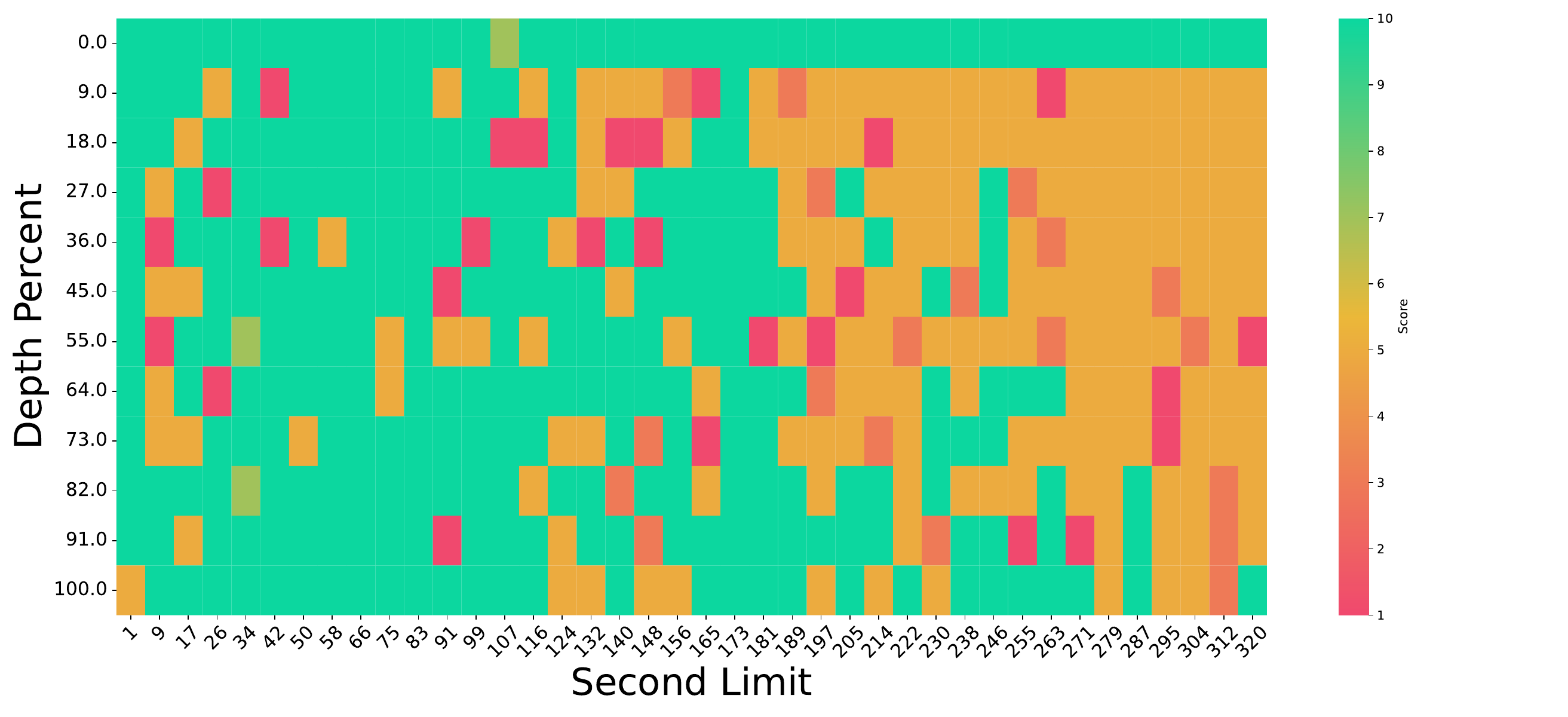} & \includegraphics[width=\imagewidth]{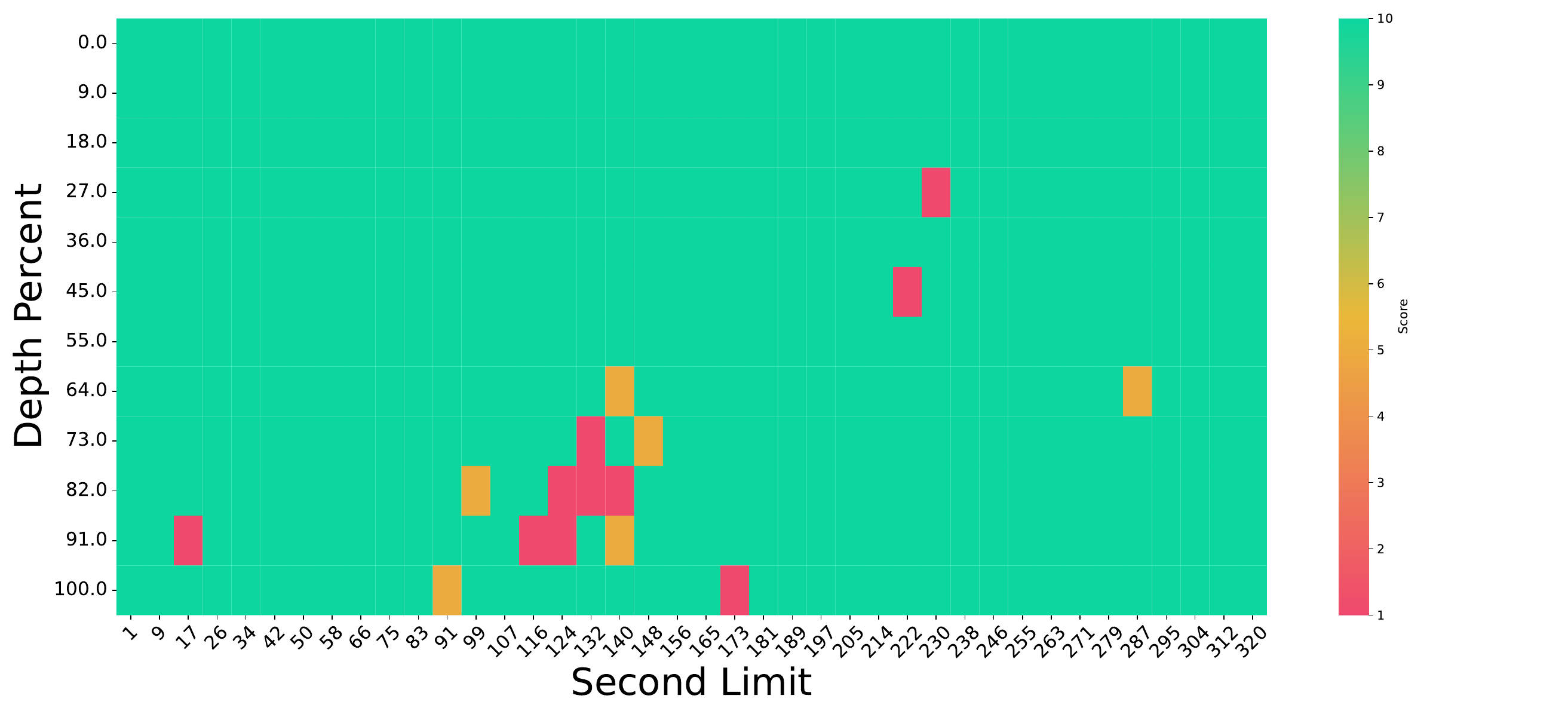} \\

    \end{tabular}
    % }
    % \begin{subfigure}{.45\textwidth}
    %     \centering
    %     \includegraphics[width=\linewidth]{figures/pllava.pdf}
    %     \caption{PLLaVA~\citep{pllava}}
    %     \label{fig:needle-pllava}
    % \end{subfigure}%
    % \hfill
    % \begin{subfigure}{.45\textwidth}
    %     \centering
    %     \includegraphics[width=\linewidth]{figures/llavanext.pdf}
    %     \caption{LLaVA-NeXT-Video-DPO~\citep{llavanext}}
    %     \label{fig:needle-llavanext}
    % \end{subfigure}%
    % \hfill
    % \begin{subfigure}{.45\textwidth}
    %     \centering
    %     \includegraphics[width=\linewidth]{figures/malmm.pdf}
    %     \caption{MA-LMM~\citep{malmm}}
    %     \label{fig:needle-malmm}
    % \end{subfigure}
    % \hfill
    % \begin{subfigure}{.45\textwidth}
    %     \centering
    %     \includegraphics[width=\linewidth]{figures/longva.pdf}
    %     \caption{LongVA~\citep{longva}}
    %     \label{fig:needle-longva}
    % \end{subfigure}
    % \hfill
    % \begin{subfigure}{.45\textwidth}
    %     \centering
    %     \includegraphics[width=\linewidth]{figures/memobridge_wort.pdf}
    %     \caption{\model w.o. retrieval}
    %     \label{fig:needle-retortv-wotr}
    % \end{subfigure}
    % \hfill
    % \begin{subfigure}{.45\textwidth}
    %     \centering
    %     \includegraphics[width=\linewidth]{figures/memobridge_mem.pdf}
    %     \caption{\model}
    %     \label{fig:needle-retortv}
    % \end{subfigure}
    % \vspace{-1em}
    \caption{\textbf{Comparison of \model with two long video understanding models on \bench~(\benchshort). }Currently, we set the context length to 320 seconds \wrt existing models' ability and set the frame rate to 1 fps to ensure the input contains the needle. The X-axis indicates the video length, and the Y-axis is the depth of the insertion point.}
    \vspace{-.1in}
    \label{fig:needle-comparison}
\end{figure*}

To address existing limitations in long-form video language understanding benchmarks, our work takes inspiration from the latest developments in the field and develops a new benchmark specifically designed for the task of identifying specific content within extensive video material, a challenge we refer to as the \bench~(\benchshort). This benchmark is unique in that it supports queries in various modalities, including text, image, and video, allowing for a more comprehensive assessment of a model's video understanding capability. 
% In \textbf{~\Cref{supp:discuss}}, we emphasize the uniqueness of our benchmark in comparison to similar ones.

% \paragraph{Haystack Setting}In our benchmark, we utilize ego-centric videos from the Ego4D~\citep{ego4d} dataset as the ``haystack''. Within this haystack, we seek to locate the ``needle'', which we provide in three distinct modalities. For the textual modality, we supply a crafted description. For the image modality, we employ DALL-E to create an image that visually represents this description. For the video modality, we use Sora to generate a short video clip based on the same description. In each case, the "needle" - whether text, image, or video - is set to a duration of 1 second. Examples of both the needle and the haystack are provided in \textbf{Appendix \ref{supp:needle}}.

% \paragraph{Benchmark Setting} To build this benchmark, we concatenate videos from Ego4D to the specified length. We then inject needles into the concatenated video at various depths and lengths. A direct question about the details within the needles is set. Afterward, we use an LLM to evaluate the response by comparing it with the ground truth and predict a consistent score from 1 to 10 with 10 indicating a perfect match. For the overall quantitative results, we calculate the average scores for additional analysis.

\paragraph{Benchmark Setting}
In \benchshort, we utilize ego-centric videos from the Ego4D~\citep{ego4d} dataset as the ``haystack,'' within which we seek to locate the ``needle'' provided in three distinct modalities: textual, image, and video. For the textual modality, we supply a crafted description; for the image modality, we use DALL-E
% \footnote{\url{https://openai.com/index/dall-e-3/}}
to create a corresponding image; and for the video modality, we use Sora~\citep{videoworldsimulators2024} generated short video clip, all based on the same description. Each ``needle'' is set to a duration of 1 second and is inserted into the concatenated Ego4D videos at various depths and lengths. To evaluate the benchmark, a direct question about the details within the needles is set, and an LLM compares the response with the ground truth, providing a score from 1 to 10, with 10 indicating a perfect match. For quantitative results, we calculate the average scores for additional analysis. \looseness=-1

% We visualize our proposed needle in a video haystack, which supports different modalities of needle, include text, image, and video. As is shown in \Cref{fig:neeld}, the needle is ``A young man is sitting on a piece of cloud in the sky, reading a book.''. For the text needle, we just append the text to the video directly; for the image and video needle, we insert the image and the video clips into the video haystack.

\noindent\underline{\textit{Comparison with similar benchmarks}}\quad{} Recent work proposes a multimodal needle-in-a-haystack benchmark MM-NIAH~\citep{wang2024needle}, which focuses on a mixture of images and documents as the haystack and only supports text and image needles. In contrast, NIVAH focuses on streaming video stacks and supports text, image, and video needles.

\paragraph{Experiment Setup} Given the limitations of current methods in understanding long videos, we designed an experiment where the ``haystack'' is a 320-second video. The ``needle'' is a 1-second video clip generated by Sora, prompted by the description, ``the young man seated on a cloud in the sky is reading a book''. The associated question posed for the experiment is, ``What is the young man seated on a cloud in the sky doing?''. We divided the context into 40 intervals and set the video depth at 12 intervals. 
% The visualization of needle and its corresponding haystack can be found in Appendix \ref{supp:needle}.

% With the advancement of video-language modeling, there has been an increasing effort to develop benchmarks for evaluating details in long videos. There are several works similar to ours that require clarification:
% MLVU~\citep{zhou2024mlvu} designs a long video question-answering (QA) dataset by inserting images into videos in advance and then annotating questions based on the videos. However, these predefined video QA datasets still suffer from potential language biases. Our method aims to pressure-test long video understanding in a flexible way, and our pipeline is not limited to specific content.
% The Gemini-1.5~\citep{gemini} technical report introduces a similar benchmark, but it only adds text to images, making it difficult to evaluate other capabilities. In comparison, our method supports different modalities of needles, including text, images, and videos. Additionally, our method is not constrained to certain frames, allowing for a fair comparison.
% MM-NIAH~\citep{wang2024needle} proposes a multimodal needle-in-a-haystack benchmark, which focuses on a mixture of images and documents as the haystack and only supports text and image needles. In contrast, our method focuses on streaming video stacks and supports text, image, and video needles.

% \subsection{Example of NIAVH}
% \label{supp:needle}
% In this section, 

\paragraph{Results and Analysis}In our experiment, we evaluate our approach with four distinct methods. These include (a) adaptive pooling~\citep{pllava}, (b) position extrapolation combined with sampling~\citep{llavanext}, (c) the integration of resampler with memory retrieval and consolidation~\citep{malmm}, and (d) video alignment with long-context LLM without compression~\citep{longva}. For each model, we standardize the video frame rate to one frame per second, aligning the number of input frames with the duration of the video in seconds. This ensures that the inputs contain  the needle information and all the models are in fair comparison. The outcomes of this evaluation are depicted in \Cref{fig:needle-comparison}.
Our analysis leads to the following key observations: 
\begin{itemize}[leftmargin=*, topsep=0pt, noitemsep]
    \item Methods utilizing an adaptive pooling strategy risk omitting crucial information, as the length of the source material (the "haystack") is often many times greater than the target segment (the ``needle''). 
    \item Pooling strategies that incorporate position extrapolation are ineffective at predicting lengths that exceed those encountered during training or fine-tuning.
    \item Combining a resampler with a retrieval strategy markedly improves the encoding of extended information in a video. However, the encoded length is ultimately constrained by the resampler's compression capacity.
    \item \model with retrieval is the most efficient at preserving previously encountered information. Nevertheless, it still exhibits shortcomings: it tends to forget earlier information and is prone to hallucination issues, such as misidentifying ``holding book'' as ``holding phone''.
\end{itemize}
% (2)  (3)  (4)  In conclusion, the findings from this experiment provide valuable insights for future research, particularly in enhancing our method to address the issue of forgetting.

\subsection{Performance Analysis}
\label{sec:analysis}

% \begin{wrapfigure}{r}{.6\linewidth}
\begin{figure}[t!]
    
    \centering
    % \vspace{-1.5em}
    \includegraphics[width=0.9\linewidth]{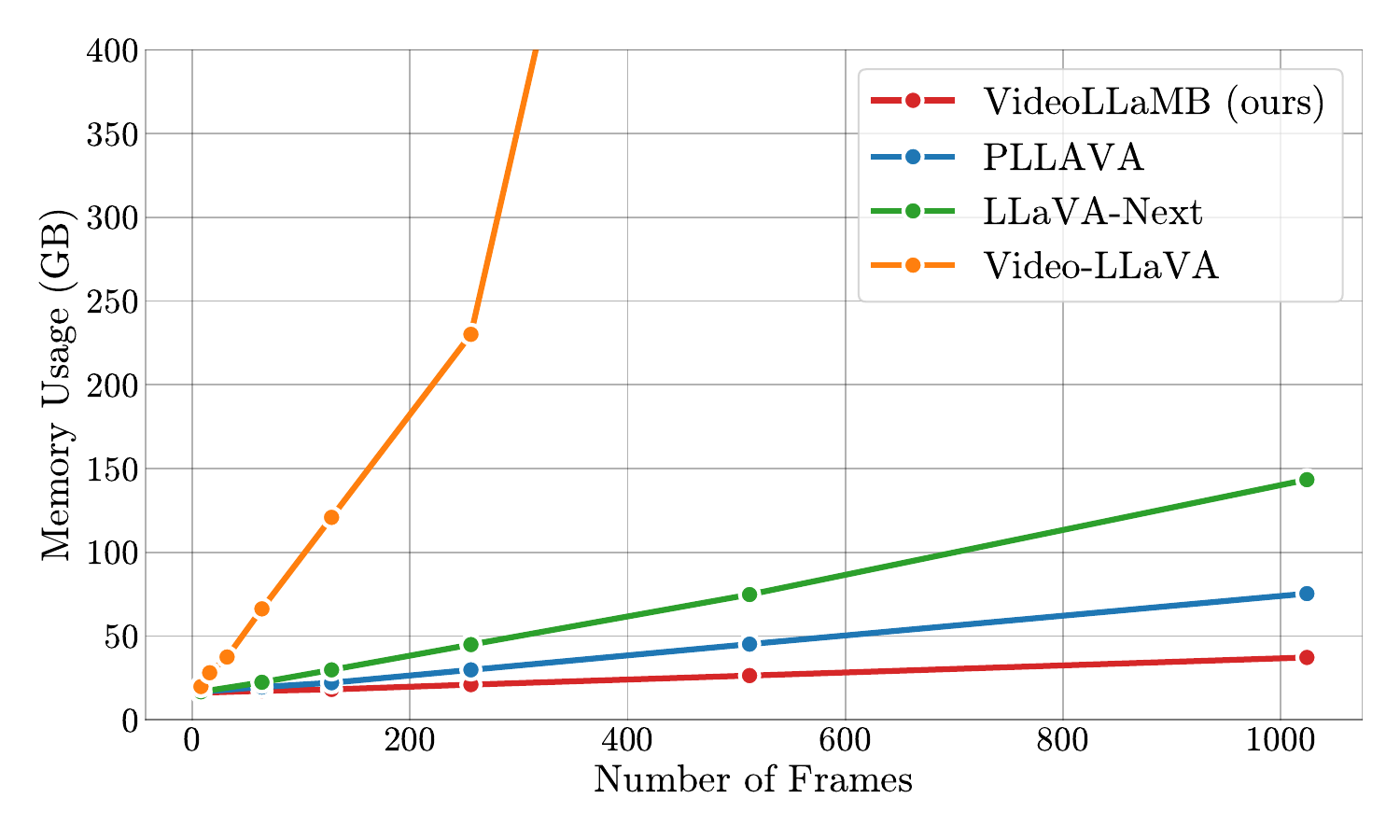}
    \vspace{-1em}
    \caption{\textbf{GPU Memory Cost.} We apply all the experiments on a single NVIDIA A800 GPU.}
    \label{fig:memory}
   \vspace{-0.2in}
% \end{wrapfigure}
\end{figure}

% \begin{figure}[ht!]
%     \centering
%     \small
%     \begin{minipage}{.47\linewidth}
%     % \vspace{-.2in}
%     \includegraphics[width=\linewidth]{figures/memory.pdf}
%     \caption{\textbf{GPU Memory Cost.} We apply all the experiments on a single NVIDIA A800 GPU.}
%     \label{fig:memory}
%     \vspace{-0.1in}
%     \end{minipage}\hfill
%     % \begin{minipage}{.47\linewidth}
%     %         \centering
%     % \includegraphics[width=\linewidth]{figures/num_frame.pdf}
%     % \caption{\textbf{Length extrapolation results} on EgoSchema dataset.}
%     % \label{fig:length}
%     %     \vspace{-0.1in}
%     % \end{minipage}
% \end{figure}

\paragraph{Memory Cost}

% \begin{wrapfigure}{r}{.5\linewidth}
%     \centering
%     \vspace{-.2in}
%     \includegraphics[width=\linewidth]{figures/memory.pdf}
%     \caption{\textbf{GPU Memory Cost.} We apply all the experiments on a single NVIDIA A800 GPU.}
%     \label{fig:memory}
%     \vspace{-0.1in}
% \end{wrapfigure}
Our model's recurrent strategy maintains a consistent visual input length to the LLM, significantly reducing GPU memory usage. While a larger memory cache theoretically requires more memory~\cite{wu2025tokenswift}, the impact is minimal due to shorter memory tokens compared to visual input tokens. The recurrent memory operates on the bridge layer, minimizing intermediate costs. In our experiments on the EgoSchema~\citep{egoschema} dataset, we compared our model against three categories: vanilla, pooling-based, and sampling-based. Results in \Cref{fig:memory} show that our methods and other fixed-length input models significantly cut memory usage, with our approach compressing input length more effectively. Our design's efficiency is evident, as the memory cache incurs negligible additional memory cost.

\paragraph{Inference Time}

Our primary concern with our approach is the potential time expenditure associated with recurrent processes and memory retrieval. 
To address this, we conducted experiments to assess the efficiency of our method in comparison to others.
The evaluation included all current methods capable of handling long videos. We tested each model on \benchshort with 300 second video cases to measure their performance for comparison.
The results \Cref{tab:efficiency}, demonstrate that our method not only outperformed the existing methods but 
% \begin{wraptable}{r}{.6\linewidth}
\begin{table}[t!]
    \centering
    \resizebox{\linewidth}{!}{%
    \begin{tabular}{lcccc}
        \toprule
         \multicolumn{1}{l}{\bf Methods} &\multicolumn{1}{c}{\bf LLM}  &\multicolumn{1}{c}{\bf Inference Time (s) $\downarrow$} &\multicolumn{1}{c}{\bf Score $\uparrow$}   \\
        \midrule
        MovieChat     & Vicuna-7B          & 143.7 & -  \\
        MALMM           & Vicuna-7B  & 14.5   & 3.39\\
        LLaVA-NeXT-Video-DPO      & Vicuna-7B        & 11.1 & 1.72 \\ 
        PLLaVA             & Vicuna-7B  & 7.4 & 1.82\\ 
        \rowcolor{LightCyan} \model (Ours)            & Vicuna-7B  & 4.21 & 5.73 \\ \bottomrule
    \end{tabular}}
    \captionof{table}{\textbf{Average Inference Time} on the 300-second videos from NIAVH. The score is the average score on NIAVH.}
    \label{tab:efficiency}
    \vspace{-0.2in}
% \end{wraptable}
\end{table}
did so even when compared to those employing a pooling strategy~\citep{pllava}. 
We attribute this improved performance to the efficient memory management mechanism integrated within the bridge layer of our method. This enables our approach to condense the visual input more effectively than others, resulting in shorter processing times of the LLM. We further analyze the composition of the inference time over videos of varying lengths, including the encoding time and generation time.

\begin{table}[t!]
    \centering
    \resizebox{\linewidth}{!}{
    \begin{tabular}{c|ccc}
    \toprule
        \textbf{Video Duration} & \textbf{Feature Process} & \textbf{Generation} & \textbf{All} \\ \midrule
        30 & 0.25 & 1.65 & 1.9 \\
        300 & 2.3 & 1.91 & 4.21 \\
        3000 & 23.4 & 8.1 & 31.5 \\
    \bottomrule
    \end{tabular}}
    \caption{\textbf{Latency Analysis (sec.)}. We evaluate the inference time of different parts on different length videos.}
    \label{tab:my_label}
\end{table}%

\subsection{Ablation Study}
\label{sec:ablate}
% ablation

% \begin{wraptable}{r}{.5\linewidth}

In this section, we present an ablation study of our method, focusing on its individual components. We analyze our method the EgoSchema dataset. The corresponding results are detailed in \Cref{tab:ablation}. Initially, we assess the effectiveness of the recurrent mechanism. To do this, we replace this mechanism with two pooling strategies: mean pooling and adaptive pooling. For comparison purposes, we configure the adaptive pooling strategies to produce a target time sequence length of 4, matching our method's settings. Our findings reveal that all pooling strategies cause a notable degradation in performance. Notably, the adaptive pooling strategy underperforms even mean pooling. We hypothesize that this discrepancy arises from differences in how training and inference are conducted; mean pooling, being more consistent, likely enhances the model's generalizability. We then evaluate the memory retrieval mechanism and observe that it is indeed capable of preserving memory to a certain degree. Lastly, we examine the impact of our semantic segmentation strategy. Compared to a uniform segmentation approach, our method is more adept at dividing videos into semantic segments. This segmentation results in a more efficient preservation of information, mitigating the information loss typically associated with sampling strategies.

% % \begin{table}[h]
% \begin{wraptable}{r}{.5\linewidth}
%     \centering
%     \resizebox{\linewidth}{!}{%
%     \begin{tabular}{c|c}
%     \toprule
%          \textbf{Compression Strategy} & \textbf{Accuracy}  \\ \midrule
%          Adaptive Pooling~\cite{pllava} & 45.6 \\
%          Token Compression~\cite{moviechat} & 42.2 \\
%          Ours & 53.8 \\
%     \bottomrule
%     \end{tabular}}
%     \caption{Comparison of different memory compression strategy}
%     \label{tab:compression}
% \end{wraptable}
% % \end{table}

% To show \model's superiority on its memory compression ability, we further compare different memory compression strategies on the same LLM, training data, and compression rate for fair comparison. We compare our method with two types trending memory compression methods: adaptive pooling ~\cite{pllava} and token compression ~\cite{moviechat}. The results on Egoschema demonstrate our method could keep the memory in a better way.

\begin{table}[t!]
    \small
    \centering
    % \vspace{-1.0em}
    \resizebox{\linewidth}{!}{%
    \begin{tabular}{lccc}
    \toprule

        \textbf{Method}  & \textbf{Accuracy} & $\Delta$ \\ \midrule

        w.o. recurrent (mean pooling)  & 51.61 & -2.19 \\
        w.o. recurrent (adaptive pooling) & 49.4 & -4.4 \\
        w.o. retrieval  & 52.2 & -1.6 \\
        w.o. segment (uniform segment)  & 52 & -1.8  \\
        w.o. mixture of images & 49.8 & -4.0 \\ 
        memory tokens only & 50.4 & -3.4 \\ 
        $k=8$ & 52.8 & -1.0 \\  \midrule
        \model (Ours)  & \textbf{53.8}  \\
         \bottomrule
         % \multicolumn{5}
    \end{tabular}
    }
    % \vspace{-1.0em}
    \caption{\textbf{Ablated results on the effects of different modules.}}
    \label{tab:ablation}
     \vspace{-1.0em}
% \end{wraptable}
\end{table}

\section{Related Work}
\label{sec:related}

\paragraph{Long Video Language Understanding}
The advancement of large language models (LLMs) has significantly improved the comprehension of long videos through their interaction with human language. Current methodologies for long video analysis are categorized into scaling-up approaches, agent-based techniques, and length extrapolation strategies. Scaling-up approaches involve increasing model parameters and expanding training datasets \citep{lwm}, or developing more efficient architectures to replace computationally intensive transformers \citep{videomamba, videomambasuit}, though these may not always be feasible. Agent-based techniques leverage LLMs' strategic planning by incorporating various visual experts for comprehensive understanding \citep{vamos, zsproc, videoagent-mem,wang2024exovip} or converting visual inputs into textual descriptions \citep{lifelongmem, llovi, doram, videoagent-cap}, but they can encounter efficiency issues and challenges with out-of-domain content. Length extrapolation extends image-language and short video-language modeling to longer durations using techniques such as temporal embeddings \citep{momentor}, prompts \citep{timechat}, position encodings \citep{lvchat, wang2024videotgb}, frame condensation \citep{moviechat}, visual token compression \citep{text-resampler, vista-llama, st-llm}, and retrieval-based methods with visual features \citep{malmm}. These often involve selective sampling, which risks information loss. Our work introduces a recurrent memory strategy to encode entire video sequences, using a memory cache to retain past memory and project the memory-augmented current semantic segment into the LLM to maintain long video understanding.

\paragraph{Anticipatory Video Planning}
Anticipatory planning, which involves predicting future actions based on past sequences and current context, has been validated as effective in language models \citep{palme, llm-planer, embodygpt}. This approach is analogous to video understanding, where action anticipation based on visual data is gaining prominence \citep{DBLP:conf/iccv/SenerY19, DBLP:conf/dagm/FarhaKSG20, DBLP:journals/jvcir/FurnariBGF17}. A growing research area is the intersection of action anticipation and goal-directed planning, enhancing AI capabilities in video understanding \citep{DBLP:conf/iccv/PatelEKIJD23, antgpt, egoplan}. This challenge is especially critical in real-time streaming environments, where systems must interpret the current state and retain an extensive memory of past events to inform decision-making. Our proposed method is well-suited to address these challenges.

\section{Conclusion}

% \model offers an advancement in video-language models by enhancing computational efficiency and efficacy. Utilizing Memory Bridge Layers with recurrent memory tokens and the SceneTiling algorithm, \model preserves crucial visual information and semantic coherence in long videos. The NIAVH benchmark robustly evaluates this capability. Empirical results show \model outperforms existing methods in long video QA, egocentric planning, and frame retrieval.
% In the future, we would like to integrate part of the LLM memory with the memory in the bridge while keeping the whole system efficient. \looseness=-1

VideoLLaMB presents a significant advancement in video-language modeling by improving both computational efficiency and long-context understanding. Through the introduction of Memory Bridge Layers with recurrent memory tokens and the SceneTiling algorithm, our approach effectively preserves essential visual cues and maintains semantic continuity across extended video sequences. 
% The NIAVH benchmark demonstrates the robustness of our framework, and 
Empirical evaluations show that VideoLLaMB consistently outperforms existing methods in tasks such as long VideoQA, egocentric planning, and frame retrieval. 
Looking ahead, we aim to explore the integration of LLM memory with the bridge memory, with a focus on preserving the system's overall efficiency.

\noindent\textbf{Acknowledgments} \quad This work was supported by the grants from the National Natural Science Foundation of China (62376031, 62372014).

\clearpage

{
    \small
    \bibliographystyle{ieeenat_fullname}
    \bibliography{main}
}

\clearpage
\appendix
% {\bf\Huge Appendices \bigskip}
% \section{Appendix}
% \label{sec:appendix}

% \DoToC

\setcounter{page}{1}
\maketitlesupplementary

\section{Parameter Analysis}
\label{supp:param}
% parameter

\begin{table}[h]
    \centering\small
    % \vspace{-1.0em}
    % \resizebox{\linewidth}{!}{%
    \begin{tabular}{ccc}
    \toprule

        \textbf{\# of Memory Tokens} & \textbf{\# of Bridge Layer}  & \textbf{Accuracy} \\ \midrule

         32 & 1  & 53.8  \\
         64 & 1 & 53\\
        32 & 3  & 54  \\
        64 & 3 & 54.6 \\
         \bottomrule
         % \multicolumn{5}
    \end{tabular}
    % }
    % \vspace{-1.0em}
    \caption{\textbf{Parameter Analysis} we apply analysis of different parameters of our framework}
    \label{tab:parameter}
    % \vspace{-.2in}
\end{table}

We conducted a detailed parameter analysis of our model, focusing on two primary aspects: the number of memory tokens and the number of bridge layers. This analysis was performed using the EgoSchema dataset, under the experimental settings in \cref{sec:setting}. The outcomes of this analysis are presented in \cref{tab:parameter}. From the results, we observed a clear trend: a simultaneous increase in the number of memory tokens and the number of bridge layers leads to a notable improvement in performance. This finding is significant as it provides valuable direction for future enhancements of our method. To optimize our model further, we propose expanding the capacity of the bridge layer by adding more parameters while concurrently exploring more efficient architectural designs.

% \section{Comprehensive Video Understanding} 
% \label{supp:mvbench}
% We also evaluate our method on a comprehensive video understanding benchmark MVBench~\citep{mvbench} to reveal our methods would improve the basic video understanding ability.
% In \cref{tab:mvbench}, we display our results on MVBench, which could reveal that our mechanism will reserve the comprehensive video understanding ability over general video understanding tasks. Notely, our method with the same training data as PLLaVA, could achieve similar performance as 13B level model. We believe our method could obtain more information whether for short or long videos.

% This is an appendix.
\section{Compression Strategy Analysis}

In this section, we further explore the memory compression ability of our metho. We compare our method with two types trending memory compression methods: adaptive pooling ~\cite{pllava} and token compression ~\cite{moviechat}. Different memory compression strategies are compared on the same LLM, training data, and compression rate for fair comparison. The results on Egoschema in \cref{tab:compression} demonstrate our method could keep the memory in a better way.

\begin{table}[h]
    \small
    \centering
    \resizebox{0.7\linewidth}{!}{
    \begin{tabular}{c|c}
    \toprule
        \textbf{Compression Strategy} & \textbf{Accuracy} \\ \midrule
        Adaptive Pooling~\cite{pllava} & 45.6 \\
        Token Compression~\cite{moviechat} & 42.2 \\
        VideoLLaMB & 53.8 \\
    \bottomrule
    \end{tabular}}
    \caption{Comparison of different memory compression strategies.}
    \label{tab:compression}
\end{table}

\section{Implementation Details}
\label{supp:implementation}

\subsection{Implementation Details}
\label{sec:setting}

In our experiment, we configured the memory tokens to a capacity of 32 and employed a single transformer layer as the bridge layer. For the training process, we set the number of training frames to 16 and limited the number of segments to 4. In order to ensure the visual encoder's plug-and-play functionality, we froze its parameters, focusing the training solely on the bridge layer and the LLMs. We utilized the Image Encoder and Video Encoder from Video-LLaVA~\citep{videollava}. In alignment with the procedures of PLLaVA~\citep{pllava}, we initialized the LLM using the LLaVA-1.5~\citep{llava15} configuration. The training dataset was identical to that used in PLLaVA, leveraging the same video data. To maintain the model's proficiency in static visual learning, we retained the fine-tuning image data from LLaVA-1.5.  Our experiments were conducted on four Nvidia A800 GPUs. Regarding other hyperparameters, we adhered to the original settings specified in the initialized models

\subsection{Parameter Details}
\label{supp:implement}
In this section, we will include more detailed implementation details. In Table \ref{tab:Hyperparam}, we demonstrate the implementation details of our method, including the details of the Bridge Layer, Retrieval Layer, and other hyperparameters of our initialized LLaVA.

\begin{table}[ht]
    \begin{center}\small
    \caption{Hyperparameters for \model.}
    \begin{tabular}{lccc}
        \toprule
        \multicolumn{1}{l}{\bf Hyperparam} &\multicolumn{1}{c}{\bf \model} \\ 
        \midrule 
        Number of Bridge Layers  & 1  \\
        Number of Retrieval Layers & 1 \\
        Bridge Layer Attention Heads & 8 \\
        Retrieval Layer Attention Heads & 8 \\
        Bridge Layer Hidden Size & 1024 \\
        Retrieval Layer Hidden Size & 1024 \\
        Vision Feature Select Layer & -2 \\
        Model Max Length & 2048 \\
        Learning Rate & 2e-4 \\
        Batch Size & 8  \\
        Epoch & 1 \\
        Warmup Ratio & 0.03 \\
        Weight Decay & 0.0 \\
        Patch Size & 14 \\
        Image Size & 224 \\
        \bottomrule
    \end{tabular}
    
    \label{tab:Hyperparam}
    \end{center}
\end{table}

\subsection{Baseline Clarification}
\label{supp:clarification}
This work miss two long-video understanding model in some benchmarks for the following reasons: (1) the MALMM is built on InstructBLIP, which limits the input query length and, therefore, can't be applied to the EgoSchma and the NExTQA benchmark. (2) MovieChat requires reloading the model at each test and requires heavy I/O pressure. Therefore, we only include the MALMM on our NIAVH benchmark for comparison. In addition, to make a fair comparison with different compression methods, we adopt these baselines in the same setting on Egoschema, and the results are illustrated on ~\cref{tab:compression}.

% \begin{table}[!ht]
%     \centering\small
%     % \resizebox{\linewidth}{!}{%
%     \begin{tabular}{lcccc}
%         \toprule
%          \multicolumn{1}{l}{\bf Methods} &\multicolumn{1}{c}{\bf LLM}  &\multicolumn{1}{c}{\bf Inference Time (s)} &\multicolumn{1}{c}{\bf Score}   \\
%         \midrule
%         MovieChat     & Vicuna-7B          & 143.7 & -  \\
%         MALMM           & Vicuna-7B  & 14.5   & 3.39\\
%         LLaVA-NeXT-Video-DPO      & Vicuna-7B        & 11.1 & 1.72 \\ 
%         PLLaVA             & Vicuna-7B  & 7.4 & 1.82\\ 
%         \model            & Vicuna-7B  & 4.21 & 5.73 \\ \bottomrule
%     \end{tabular}
%     % }
%     \captionof{table}{Average Inference Time on the 300-second videos from NIAVH. The score is the average score on NIAVH. The MovieChat can't be applied to complete the pressure test for the runtime reason}. 
%     \label{tab:efficiency-all}
% \end{table}

% \begin{figure*}
%     \centering
%     \includegraphics[width=\textwidth]{figures/streaming1.pdf}
%     \caption{Qualitative results on streaming dense caption tasks. The video is randomly selected from the NExTQA validation set. Our method could accurately recognize the camera change and zoom out, and predict the corresponding captions.}
%     \label{fig:streaming}
% \end{figure*}

\section{Qualitative Results}

\paragraph{Planning}
\begin{figure}[thbp!]
    \centering
    \includegraphics[width=\linewidth]{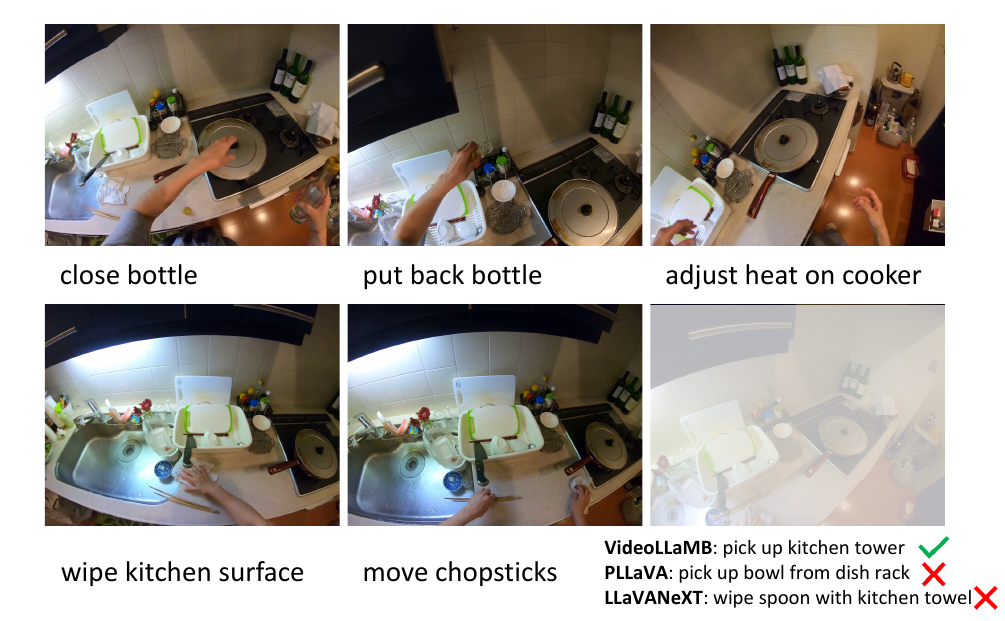}
    % \vspace*{-17pt}
    \captionof{figure}{\small \textbf{Qualitative results on EgoPlan.}}
    \label{fig:case-egoplan}
\end{figure}

We present the qualitative outcomes of various approaches on EgoPlan, as depicted in the \cref{fig:case-egoplan}. The target goal is ``\texttt{clean and organize kitchen}''.  Our method showcases effective reasoning based on previous steps and the current state, in contrast to other methods that tend to make predictions based solely on the initial or final visual inputs. Consequently, our approach enhances the model's capability in planning tasks.

\begin{figure*}[thbp!]
    \centering
    \includegraphics[width=\textwidth]{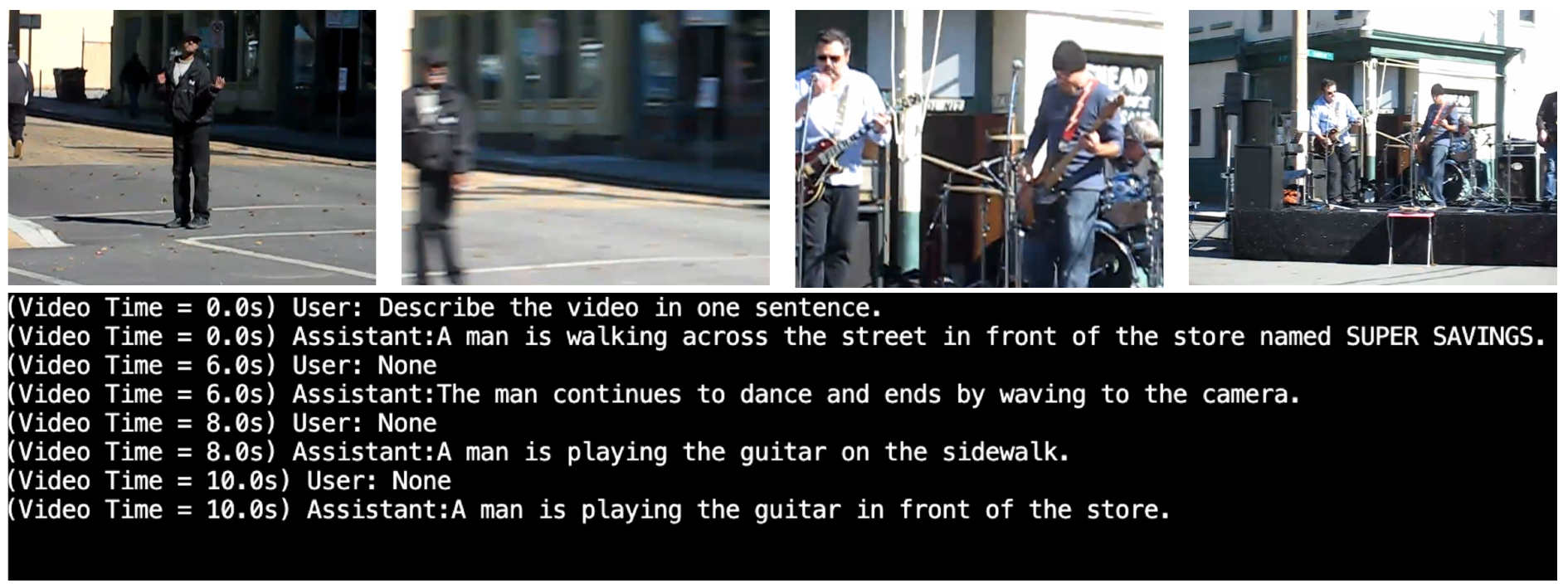}
    \caption{\textbf{Qualitative results on streaming dense caption tasks.} The video is randomly selected from the NExTQA validation set. Our method could accurately recognize the camera change and zoom out, and predict the corresponding captions.}
    \label{fig:streaming}
\end{figure*}

\paragraph{Streaming Caption}
In Figure~\cref{fig:streaming}, we present the qualitative results of the streaming caption task. At the commencement of the video, the model is provided with the query: ``Describe the video in one sentence''. Subsequently, at timestamps $0.0$ seconds, $6.0$ seconds, $8.0$ seconds, and $10.0$ seconds, the model autonomously generates captions in response to changes in the video scene, without requiring any user input.

\subsection{Example of NIAVH}
\label{supp:needle}
\begin{figure*} [thbp!]
    \centering
    \begin{tabular}{cc}
         \textbf{Needle:}  \textit{A young man is sitting on a piece of cloud in the sky, reading a book.} \\
             \includegraphics[width=\textwidth]{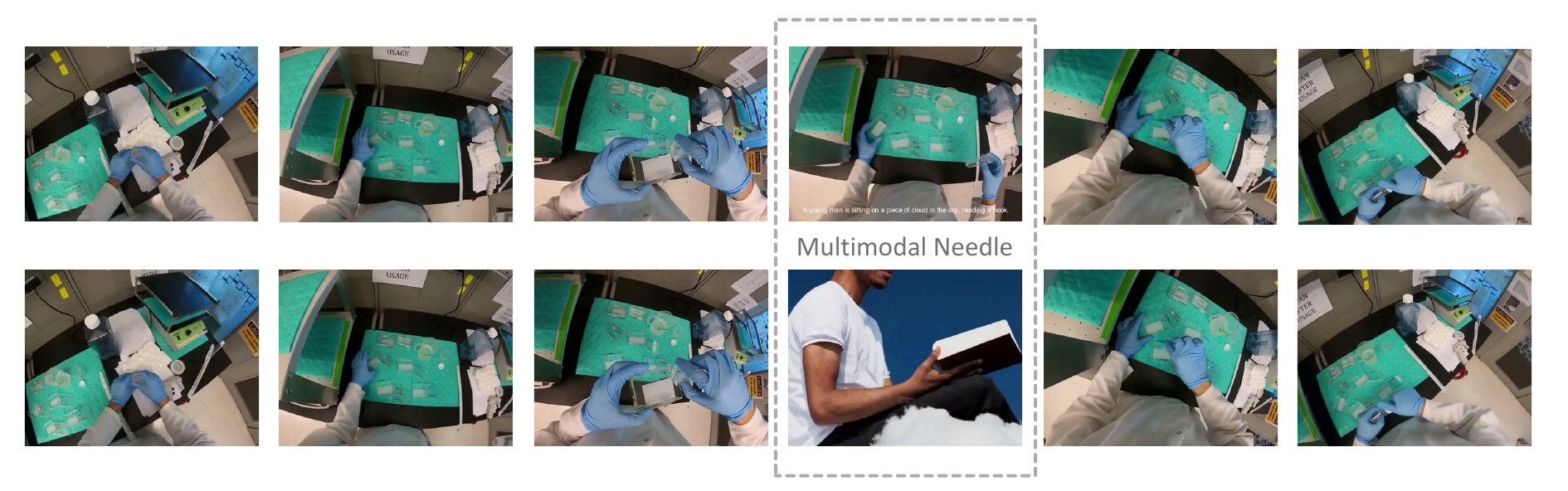}
    \end{tabular}
    \caption{\textbf{Example of \benchshort.} For the text needle, the description is appended directly to the video; for the image and video needles, the corresponding image and video clips are inserted into the video haystack.}
    \label{fig:neeld}
\end{figure*}

In this section, we visualize our proposed needle in a video haystack, which supports different modalities of needle, include text, image, and video. As is shown in \cref{fig:neeld}, the needle is ``A young man is sitting on a piece of cloud in the sky, reading a book.''. For the text needle, we just append the text to the video directly; for the image and video needle, we insert the image and the video clips into the video haystack.

% \begin{figure*}
%     \centering
%     \includegraphics[width=\textwidth]{figures/case_needle.pdf}
%     \caption{Example of our needle in a video haystack benchmark.}
%     \label{fig:neeld}
% \end{figure*}

\end{document}